\documentclass{article}
\pdfoutput=1

\usepackage{PRIMEarxiv}

\usepackage[utf8]{inputenc} 
\usepackage[T1]{fontenc}    
\usepackage{hyperref}       
\usepackage{url}            
\usepackage{booktabs}       
\usepackage{amsfonts}       
\usepackage{nicefrac}       
\usepackage{microtype}      
\usepackage{lipsum}
\usepackage{fancyhdr}       
\usepackage{graphicx}       
\usepackage{multirow}
\usepackage{array}
\usepackage{amsmath}
\graphicspath{{media/}}     

\newcommand*\rot{\rotatebox{90}}
\title{Spatiotemporal-Linear: Towards Universal Multivariate Time Series Forecasting
}

\author{
  Aiyinsi Zuo \\
  University of Rochester \\
  \texttt{azuo@u.rochester.edu} \\
   \And
     Haixi Zhang \\
  University of Rochester \\
  \texttt{hzh104@u.rochester.edu} \\
   \And
  Zirui Li \\
  University of Rochester \\
  \texttt{zli133@u.rochester.edu}
     \And
    Ce Zheng \\
    University of Central Florida \\
  \texttt{cezheng@knights.ucf.edu}
}

\begin{document}

\maketitle

\begin{abstract}
Within the field of complicated multivariate time series forecasting (TSF), popular techniques frequently rely on intricate deep learning architectures, ranging from transformer-based designs to recurrent neural networks. However, recent findings suggest that simple Linear models can surpass sophisticated constructs on diverse datasets. These models directly map observation to multiple future time steps, thereby minimizing error accumulation in iterative multi-step prediction. Yet, these models fail to incorporate spatial and temporal information within the data, which is critical for capturing patterns and dependencies that drive insightful predictions. This oversight often leads to performance bottlenecks, especially under specific sequence lengths and dataset conditions, preventing their universal application. In response, we introduce the SpatioTemporal-Linear (STL) framework. STL seamlessly integrates time-embedded and spatially-informed bypasses to augment the Linear-based architecture. These extra routes offer a more robust and refined regression to the data, particularly when the amount of observation is limited and the capacity of simple linear layers to capture dependencies declines. Empirical evidence highlights STL's prowess, outpacing both Linear and Transformer benchmarks across varied observation and prediction durations and datasets. Such robustness accentuates its suitability across a spectrum of applications, including but not limited to, traffic trajectory and rare disease progression forecasting. Through this discourse, we not only validate the STL's distinctive capacities to become a more general paradigm in multivariate time-series prediction using deep-learning techniques but also stress the need to tackle data-scarce prediction scenarios for universal application. Code will be made available.
\end{abstract}


\section{Introduction}
Time series forecasting (TSF), a subject of immense importance that transcends various domains, has garnered considerable research interest—from financial markets \cite{chen2003application} and electric utilities \cite{zhou2021informer} to network traffic \cite{lopez2017network}, autonomous driving \cite{kotseruba2016joint}, and environmental forecasting \cite{perez2006integrated}. Though data in individual time series vary, research suggested that most series possess inherent periodicity and spatial dependency\cite{cressie2015statistics}. With the development of deep learning techniques in recent years, recurrent neural networks (RNN) \cite{medsker2001recurrent} and temporal convolution networks (TCN) \cite{lea2017temporal} consistently outperform traditional models like AR (autoregressive) and MA (moving-average) \cite{mahmoud2021survey} by deciphering such relationships. 

\begin{figure}[t]
    \centering
    \includegraphics[width=1\columnwidth]{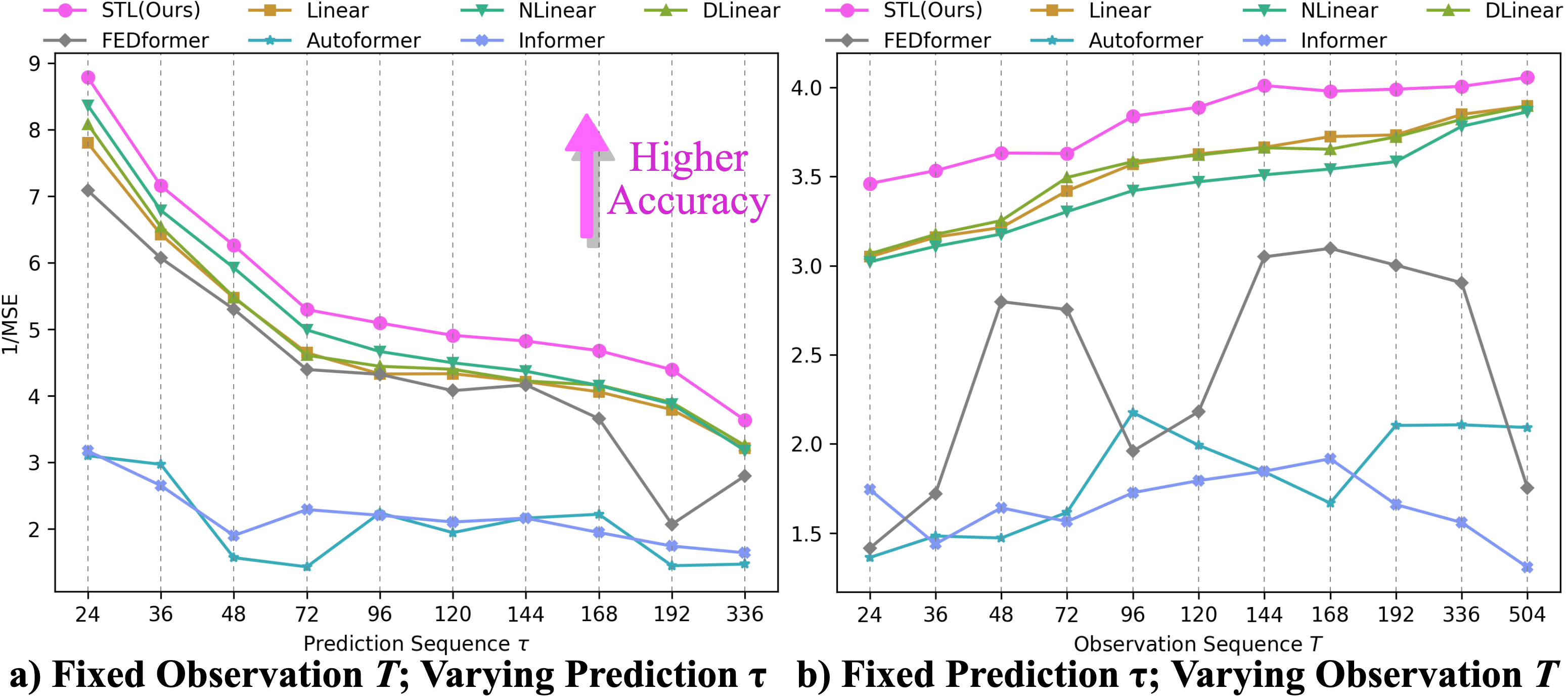}
    \caption{Illustration of STL's superiority on \textbf{various prediction lengths} given same observation sequence of 48 (left) and information richness given \textbf{same prediction length} of 336 (right); both use Weather Dataset. Zoom in for detail.}
    \label{fig:enter-label}
\vspace{-5pt}
\end{figure}
Recently, the success of transformers in natural language processing has inspired their application to time series forecasting \cite{wen2022transformers}, yielding notable enhancements over RNNs by harnessing both temporal information and spatial dependencies among variables \cite{grigsby2021long,lim2021temporal}. However, emerging research suggests that transformer-based models, similar to RNNs, follow an iterative multi-step prediction approach and, thus, inevitably suffer from error accumulation effects. To solve this issue, they proposed LTSF Linear  \cite{zeng2023transformers}, which is a set of linear-based models that directly predict the entire future sequence of one variable based on only its known observation (univariate), attaining superior performance over transformers on considerable occasions. 

Although LTSF Linear performs exceptionally well in long-term sequence prediction with sufficient information, it overlooks two other important scenarios in time series forecasting: short-term prediction and scarce data input. The corresponding applications include high-frequency trading in the stock market and traffic trajectory prediction in autonomous driving. In our investigation, these linear models suffer from significant performance drops when the input sequence (historical data) is short, faltering behind transformer-based counterparts$^1$.

To address the limitation for applications under more scenarios, we propose a SpatioTemporal-Linear (STL) framework that incorporates spatial and temporal information, both critical in multivariate time series. Spatial information captures the interrelationships between different variables, and temporal intricacies (sequential and date-time stamps) can offer a more nuanced understanding of observation and imply future trends. By effectively modeling them, we enable STL to refine and even rectify predictions for more robust regression than LTSF Linear \footnote{theoretical proof in Methodology, empirical in Experiments}. Specifically in structure, STL features three pathways: linear-based core, time-embedded temporal, and dependency-guided spatial routes. These pathways collaboratively process a given multivariate observation sequence, integrating to produce refined future predictions.

Our empirical evaluations highlight STL's distinct edge. It consistently outperforms benchmark models on datasets from \cite{zeng2023transformers} under diverse prediction durations and observation lengths simulating different prediction scenarios. An ablation study further illuminates the efficacy of the routes. Expanding our validation, we incorporated a traffic trajectory dataset from \cite{kotseruba2016joint}: under these data-constrained environments, STL reaffirms its supremacy.

The STL framework thus offers three main innovations:

\textbf{A Novel Spatiotemporal Linear-Based Framework:} We present an innovative linear-based model uniquely designed to harness both temporal and spatial information, setting a new paradigm in multivariate time series forecasting.

\textbf{Versatility in Prediction Lengths:} Demonstrating superiority in performance, our approach consistently outperforms prevailing Linear and Transformer benchmarks, regardless of whether the prediction lengths are long or short, visualized in Fig.1(a). This exhibits the model's adaptability to a wide spectrum of forecasting scenarios.

\textbf{Highlights on Varied Information Richness:} A standout feature of our model is its resilience against data scarcity. Whether faced with scarce or dense input data, our model maintains unswerving predictive accuracy (Fig.1(b)). To the best of our knowledge, it is the \textbf{first work} to investigate the deficiency in LTSF-Linear's performance during data-scarce scenarios and offer a solution for more universal applications.

\section{Generalized TSF Problem Definition}
Given a time series forecasting challenge encompassing \(C\) variables over \(T\) observations, with an objective to predict \(\tau\) future steps, we frame this as leveraging a matrix \(X_{1:T} \in \mathbb{R}^{T\times C}\) to project matrix \(X_{T+1:T+\tau} \in \mathbb{R}^{\tau\times C}\).  Note, sequence can start at any step, i.e. $X_{t_0:t_0+T}$, but to simplify theoretical derivations, $X_{1:T}$ is used in the rest of the article. Adapted from \cite{cressie2015statistics}, the prediction is articulated as
\begin{equation}
    X_{T+1:T+\tau} = X_{1:T}\beta(s,t) + w(s,t) + \epsilon(s,t) \notag
\end{equation}
Here, \(X_{1:T}\beta(s,t)\) denotes a spatiotemporal regression; The term \(w(s,t)\) encapsulates unexplained variability with spatial and temporal intricacies---for example, a synchronous movement between the London and Paris stock exchanges in response to unseen regional factors; The component \(\epsilon(s,t)\) represents noise devoid of spatiotemporal pattern.

Furthermore, we spotlight two pivotal scenarios often overshadowed: data overwhelming and data scarce. In the former, the observational span vastly exceeds the prediction horizon, represented as \(T\gg\tau\), with high-frequency trading serving as an example where vast historical data informs split-second price movement predictions \cite{jones2013we}. Conversely, the latter scenario, where the observational duration is significantly truncated compared to the predictive horizon, exemplified as \(T\ll\tau\), is typical in autonomous driving. Here, ego vehicles immediately forecast the future actions of surrounding traffic upon detection \cite{rasouli2019pie}.

\begin{figure}[t]
\centering
\includegraphics[width=0.6\columnwidth]{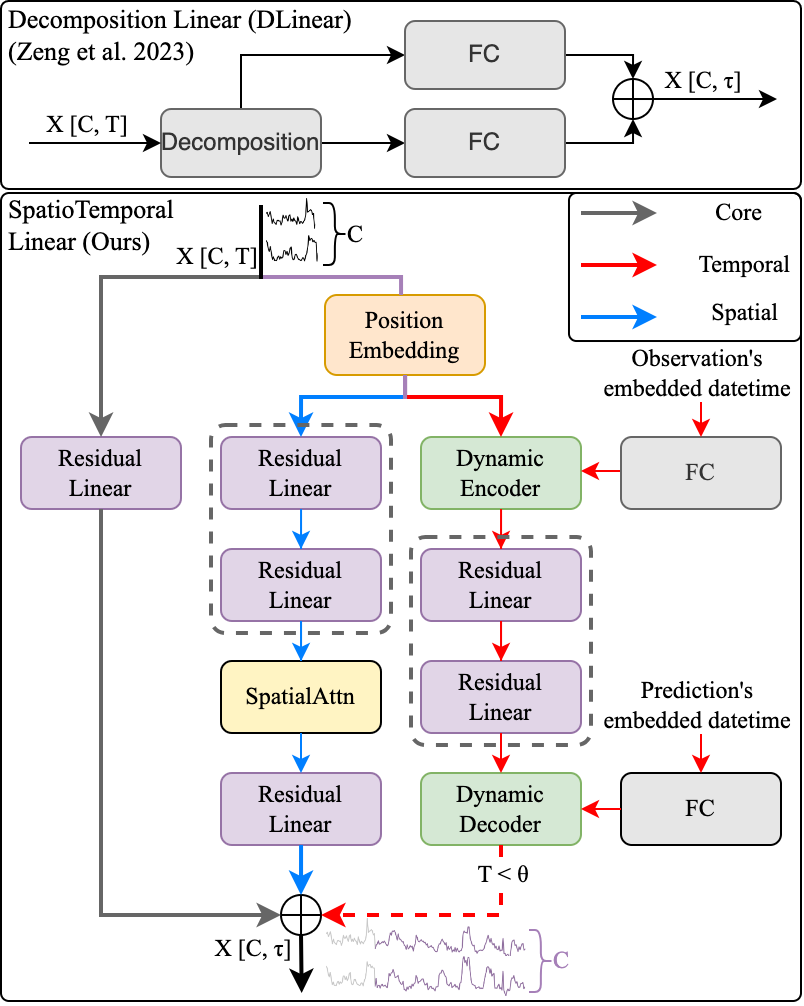} 
\caption{\textbf{Architecture Blueprint:} we visually compare the structure of D-Linear, proposed by \cite{zeng2023transformers} (top) and the three-route design of STL (bottom). Blocks in dashed boxes represent a dual Res-L encoder-decoder structure.}
\label{fig1}
\end{figure}

\begin{figure*}[t]
\centering
\includegraphics[width=1\columnwidth]{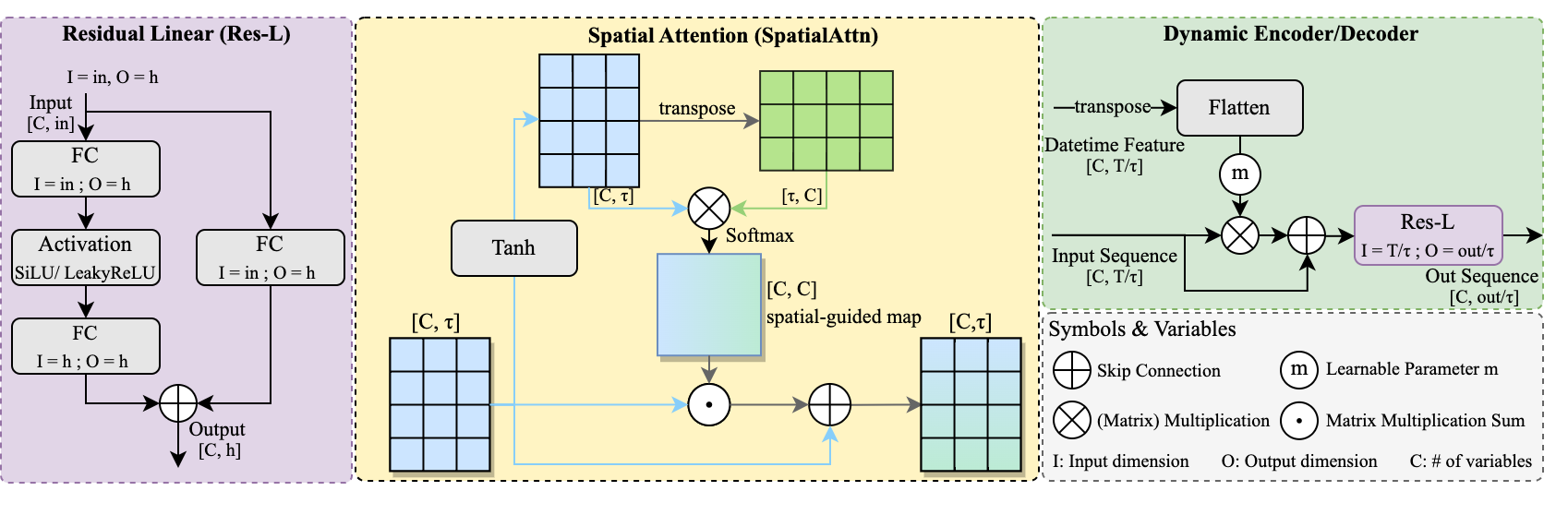} 
\caption{\textbf{Modules in STL:} We selectively illustrate the structures of Residual, Spatial Attention, and Dynamic Encoder/Decoder, the core innovations in Core, Spatial, and Temporal routes, respectively.}
\label{fig1}
\end{figure*}
\section{Related Works}
\subsection{Transformer Based Models}
Transformers, originating from \cite{vaswani2017attention} work, have been groundbreaking in areas such as natural language processing \cite{devlin2018bert} and computer vision\cite{han2022survey}. Their success is attributed to their ability to jointly attend to different types of features efficiently using their multi-head self-attention mechanism\cite{vaswani2017attention}. This strength has spurred interest in their application to time series modeling.

Time series data is intrinsically temporal, but also often exhibits spatial patterns, especially in multivariate time series where multiple series interact \cite{lim2021time}. Recognizing this, researchers have proposed Transformer adaptations specifically for multivariate TSF, constantly outperforming previous methods of GRU\cite{chung2014empirical}, LSTM\cite{gers2000learning}, ARIMA\cite{nelson1998time}, and more.

For instance, Pyraformer \cite{liu2021pyraformer} and FEDformer \cite{zhou2022fedformer} introduce mechanisms that delve into different resolutions and frequencies, capturing both short-term and long-term temporal and spatial dependencies. Additionally, Autoformer \cite{wu2021autoformer} and LogTrans \cite{li2019enhancing} enhance the way temporal features are incorporated using learnable timestamps and temporal convolution layer, respectively. \cite{xu2020spatial, you2022spatiotemporal,grigsby2021long} have proposed models that consider both the chronological order of data points (temporal information) and the interactions or relationships between different series or features (spatial information).

However, as demonstrated by \cite{zeng2023transformers}, transformers, despite utilizing positional encoding and token-based embeddings for sub-series to retain ordering details, still suffer from error accumulation due to iterative prediction approach and information loss due to the permutation-invariant self-attention mechanism.

\subsection{Linear Models}
Recent research by \cite{zeng2023transformers} unveiled three distinct linear models that surpass their transformer counterparts: Linear, DLinear, and NLinear. At its core, the Linear models utilize only historical time series to anticipate each future event through a calculated weighted sum operation, minimizing error accumulation effects.

Yet, an inherent drawback mars these models. Fundamentally anchored in a univariate linear layer, they employ univariate observations $x_{1:T}\in \mathbb{R}^{T\times 1}$ to prognosticate another univariate sequence $x_{T+1:T+\tau}\in \mathbb{R}^{\tau \times 1}$ for each variable in the multivariate sequence. This approach overlooks critical aspects: the potential impact of external variables leads to a spatial information loss, while neglecting specific sequential and date-time properties results in a truncated temporal perspective. Consequently, the regression coefficients extracted by these linear models are fragmentary, represented by $\beta(s',t')|s' \subset s, t' \subset t$.\footnote{Refer to Appendix C for proof supporting this claim} Moreover, data in \cite{zeng2023transformers} shows sensitivity to data scarcity, exhibiting performance deficiency as observation gets increasingly shorter.
\subsection{Spatiotemporal Application}
Spatiotemporal information refers to data that captures both the spatial location and temporal duration of phenomena. It integrates two critical dimensions: 'where' (space) and 'when' (time). In machine learning, harnessing spatiotemporal information has been proven to lead to more robust and sophisticated models. 

For instance, in the domain of Human Pose Estimation, a spatial-temporal transformer was designed to accurately model human joint relations within each frame and temporal correlations across frames\cite{zhao2023poseformerv2,hua2023part}; similarly, in the domain of Video Processing and Analysis, the concept of multiscale spatiotemporal tracking has been explored to enhance situation awareness.\cite{hampapur2005smart}.

In time series forecasting, spatiotemporal data is more pervasively used. It has proven to be conducive to many methods, including graph neural networks (GNNs)\cite{karimi2021spatiotemporal}, temporal convolution networks (TCNs)\cite{fan2023parallel}, and previously mentioned transformers.

\section{Methodology}
To collectively utilize critical spatiotemporal information, we employ a three-route approach (core, temporal, and spatial route), as shown in Fig.2. Each of the three pathways ingests an observational sequence $X_{1:T} \in \mathbb{R}^{T\times C}$, and in turn, produces a forecasted sequence $X_{T+1:T+\tau} \in \mathbb{R}^{\tau\times C}$. As the output, instead of processing each route in isolation, they are aggregated through summation, employing a skip connection technique. This aggregation consolidates unique features from each route while offering refinement effects, aiming to provide comprehensive and robust predictions that encapsulate both spatial and temporal attributes, or:
\begin{equation}
X_{T+1:T+\tau}=X^{core}_{T+1:T+\tau}+X^{temp}_{T+1:T+\tau}+X^{spat}_{T+1:T+\tau}\notag
\end{equation}
The design specifics of each route are illustrated below and essential innovative modules are visualized in Fig.3.

\subsection{Core Route}
In the design of the core route, we emphasize both transformation and preservation. To this end, we introduce the ``Residual Linear'' (Res-L) layer to achieve such objectives.

This layer comprises two subroutes: a dual linear transformation and a direct linear skip connection. The dual linear transformation processes the sequence \(X_{1:T}\) through a stack of linear layers punctuated by an activation function to provide the architecture with non-linearity. The initial layer accepts an observation length \(T\) and projects it to a hidden space with dimensions equivalent to the prediction length, denoted as \(h=\tau\). Following this, a subsequent layer decodes the intermediate representation, maintaining the dimensionality \(\tau\) as the output. Parallelly, the skip connection maps from dimension \(T\) directly to \(h = \tau\) to attain the output. Consequently, the output from the Res-L layer (i.e. core route) can be expressed as 
\begin{equation}
X_{T+1:T+\tau}^{core} = L_1(X_{1:T})+L_3\circ g(L_2(X_{1:T}))\notag
\end{equation}
where \(g\) symbolizes the activation function---a choice between SiLU and LeakyReLU, \(\circ\) represents function composition, and $L_{1,2,3}$ refers to simple linear layers within the residual linear block. Notably, dropouts are applied after $L_3$ depending on the datasets. 

This design ensures two critical features. First, the addition of linearly derived $X^{L_1}_{T+1:T+\tau}$ to the non-linearly transformed output helps maintain a direct linkage to the original data, mitigating potential information loss during transformation. Second, this residual connection introduces an inherent regularization effect \cite{he2016deep}. The underlying principle is that learning residuals often streamline the optimization process, offering both computational efficiency and enhanced stability.

In summary, this core route serves as a fast-track conduit that provides a baseline around which more intricate spatiotemporal patterns can be learned through additional routes.
\subsection{Temporal Route}
The temporal route is tailored to process sequential and date-time data and emerges as a crucial addition within our broader model architecture.

For the entire route, we employ a coarse-refined strategy: initially, the input $X_{1:T}$ undergoes positional embedding (PE) to capture its temporal structure, resulting in $X^{pe}_{1:T}$. Then we implement a dynamic encoder (DE) that processes this embedded input to a hidden dimension space $h=hidden\_size$. This output is then traversed through two "Residual Linear" (Res-L) layers (dual Res-L encoder-decoder structure) to form a preliminary prediction $X^{prel}_{T+1:T+\tau}$. Culminating this process, a dynamic decoder (DD) refines the preliminary decoding to generate a temporal-guided prediction:
\begin{equation}
    X^{temp}_{T+1:T+\tau}= \text{DD} \circ \text{Res-L}\circ\text{Res-L} \circ \text{DE}(\text{PE}(X_{1:T}))\notag
\end{equation}

\textbf{Positional Embedding:} Regular linear layers do not explicitly consider sequentiality within observation sequences. To counteract this limitation, we've integrated a ``PositionalEncodingLayer'' (PE) that augments input data using sinusoidal positional encodings---this approach is inspired by \cite{vaswani2017attention}. For a sequence length \( T \) and \( C \) variables, positional encodings for each time step, \( p \), are computed as:
\begin{align}
\text{PE}(p, 2i) &= \sin\left(p \times \frac{1}{10000^{\frac{2i}{C}}}\right) \notag\\
\text{PE}(p, 2i+1) &= \cos\left(p \times \frac{1}{10000^{\frac{2i}{C}}}\right)\notag
\end{align}
Here, $i$ is an integer where \(0 \leq i \leq \left\lfloor \max\left(\frac{C}{2}, \frac{C-1}{2}\right) \right\rfloor\). These encodings, when added during the forward pass, introduce channel-wise sequential context as well as channel-specific spatial context without modifying original data characteristics.

\textbf{Date-Time Embedding:} After considering sequential information from the positional embedding, we employ a ``DateTimeEmbedding" layer that embeds additional date-time information to the observation and the decoding sequence. 

Standard linear models tend to overlook the cyclical and categorical essence of date-time elements, especially in short observation sequences containing \textit{incomplete periodicity knowledge}. Our ``DateTimeEmbedding'' module transforms discrete date-time components into dense vector representations. Each component---like  dates, weekdays, or hours---is mapped to a continuous space through embedding layers with respective numbers of unique embedding indices (31, 7, 24). These embeddings are then concatenated and processed to attain machine-understandable vectors and consistent dimensionality through a linear reducer:
\begin{equation}
\text{datetime\_feature}\in\mathbb{R}^{T\times 1} = L_{reducer}([e_{\text{date}}; e_{\text{weekday}}; e_{\text{hour}}])\notag
\end{equation}

\textbf{Dynamic Encoder/Decoder:} To use the date-time information effectively, we implement Dynamic Encoder (DE) and Dynamic Decoder (DD), which possess the same structure to modulate the individual effects of these embeddings. Both layers introduce a learnable gating mechanism, governed by parameter $m$, to control the degree of influence the date-time embeddings exert on the positional embedded $X^{pe}_{1:T}$. After modulating the date-time embeddings with $m$ and normalizing, they're combined with $X^{pe}_{1:T}$: 
\begin{equation}
X' = X^{pe}_{1:T} + m \times \text{normalize}(\text{datetime features})\notag
\end{equation}
where,\vspace{-1em}
\begin{equation}
    \text{normalize}(x) = \frac{x-min(x)}{max(x)-min(x)}\notag
\end{equation}
Subsequently, $X'$ is processed using a Residual Linear (Res-L) layer to generate the output of the DE/DD, whose dimension is $h=hidden\_size/ \tau$, respectively. This design ensures that the model learns the optimal extent to which date-time information should be integrated, rather than imposing a fixed integration strategy.

Lastly, our model only merges the temporal route's output when the observation sequence length falls below a specific threshold $\theta_{T}$. This threshold is indicative of periodicity in the observation sequence. In our exploration: As the sequence length amplifies, inherent periodicity $I(X)$ becomes evident. Thus, overlaying this with $I(d)$ to supplement cyclical context may usher redundancy. For instance, in stock prediction, day-of-week information becomes redundant in models trained on year-long datasets that inherently account for weekly trading patterns. Overemphasizing known cyclical patterns might cause the model to fit noise, deteriorating its performance on unseen data. Hence, we opt to incorporate date-time embeddings to improve the robustness of the model only when periodicity in the data is not explicit $T \leq \theta_T$.
\begin{table*}
\centering
{\fontsize{8pt}{10pt}\selectfont
\newcolumntype{C}[1]{>{\centering\arraybackslash}p{#1}}
\begin{tabular}{C{0.4cm}C{0.4cm}C{0.4cm}|*{2}{C{0.6cm}}||*{6}{C{0.5cm}}||*{6}{C{0.5cm}}}
\hline
\multicolumn{3}{c|}{Methods}  & \multicolumn{2}{c||}{\textbf{STL(Ours)}} & \multicolumn{2}{c|}{\textbf{Linear}} & \multicolumn{2}{c|}{\textbf{DLinear}} & \multicolumn{2}{c||}{\textbf{NLinear}} & \multicolumn{2}{c|}{\textbf{Autoformer}}  & \multicolumn{2}{c|}{\textbf{Informer}} & \multicolumn{2}{c}{\textbf{FEDformer}} \\
\hline
Data & $T$ & $\tau$  & MSE & MAE & MSE & MAE & MSE & MAE & MSE & MAE & MSE & MAE & MSE  & MAE & MSE & MAE \\ 
 \hline
\multirow{5}{0.4cm}{\rot{Electricity}} & \multirow{5}{*}{$T_{best}$} & 24  & \textbf{0.094} & \textbf{0.192} & \underline{0.106} & \underline{0.205} & \underline{0.106} & \underline{0.205} & 0.107 & 0.206 & 0.174 & 0.296 & 0.284 & 0.375 & 0.158 & 0.279\\
& & 48  & \textbf{0.110} & \textbf{0.208} & \underline{0.121} & 0.220 & \underline{0.121} & \underline{0.219} & 0.122 & 0.220 & 0.186 & 0.303 & 0.297 & 0.386 & 0.175 & 0.291 \\
& & 96  & \textbf{0.129} & \textbf{0.225} & \underline{0.136} & 0.234 & \underline{0.136} & \underline{0.233} & 0.138 & 0.234 & 0.201 & 0.315 & 0.311 & 0.396 & 0.188 & 0.304 \\
& & 192  & \textbf{0.150} & \textbf{0.247} & \textbf{0.150} & \textbf{0.247} & \textbf{0.150} & \textbf{0.247} & \underline{0.152} & \underline{0.248} & 0.225 & 0.341 & 0.384 & 0.445 & 0.197 & 0.311 \\
& & 336  & \textbf{0.163} & \textbf{0.260} & \underline{0.165} & 0.264 & \underline{0.165} & 0.265 & 0.168 & \underline{0.263} & 0.233 & 0.344 & 0.372 & 0.441 & 0.212 & 0.327 \\
\hline
\multirow{5}{*}{\rot{Etth1}} & \multirow{5}{*}{$T_{best}$} & 24 & \textbf{0.308} & \textbf{0.351} & 0.320 & 0.368 & \underline{0.313} & \underline{0.362} & 0.317 & 0.366 & 0.392 & 0.424 & 0.518 & 0.512 & 0.330 & 0.387 \\
& & 48  & \textbf{0.340} & \textbf{0.376} & 0.344 & 0.381 & \textbf{0.340} & \textbf{0.376} & \underline{0.342} & \underline{0.380} & 0.414 & 0.445 & 0.598 & 0.530 & \underline{0.342} & 0.389 \\
& & 96  & \textbf{0.370} & \textbf{0.394} & 0.373 & 0.398 & \textbf{0.370} & \textbf{0.394} & \underline{0.371} & \underline{0.397} & 0.462 & 0.469 & 0.752 & 0.611 & 0.381 & 0.421 \\
& & 192  & \textbf{0.404} & \textbf{0.413} & 0.409 & 0.422 & \textbf{0.404} & \underline{0.416} & \underline{0.406} & 0.417 & 0.509 & 0.507 & 0.979 & 0.748 & 0.429 & 0.455 \\
& & 336  & \textbf{0.433} & \textbf{0.433} & 0.439 & 0.442 & \underline{0.434} & \underline{0.434} & 0.435 & 0.436 & 0.526 & 0.529 & 1.040 & 0.786 & 0.443 & 0.462 \\
\hline
\multirow{5}{*}{\rot{Ettm1}} & \multirow{5}{*}{$T_{best}$} & 24  & \textbf{0.194} & \textbf{0.281} & \underline{0.211} & \underline{0.284} & 0.215 & 0.286 & 0.216 & 0.287 & 0.331 & 0.401 & 0.349 & 0.404 & 0.279 & 0.352 \\
& & 48  & \textbf{0.262} & \textbf{0.324} & \underline{0.271} & \underline{0.326} & 0.276 & 0.328 & 0.279 & 0.331 & 0.389 & 0.432 & 0.479 & 0.489 & 0.344 & 0.394 \\
& & 96  & \textbf{0.291} & \textbf{0.342} & \underline{0.300} & \underline{0.343} & 0.309 & 0.355 & 0.305 & 0.347 & 0.456 & 0.482 & 0.549 & 0.530 & 0.350 & 0.410 \\
& & 192 & \textbf{0.330} & \underline{0.366} & \underline{0.335} & \textbf{0.364} & \underline{0.335} & \underline{0.366} & 0.341 & 0.369 & 0.427 & 0.461 & 0.702 & 0.602 & 0.379 & 0.425 \\
& & 336  & \underline{0.367} & 0.395 & 0.368 & \underline{0.386} & \textbf{0.366} & \textbf{0.384} & \underline{0.367} & \textbf{0.384} & 0.493 & 0.507 & 0.912 & 0.698 & 0.415 & 0.443 \\
\hline
\multirow{5}{*}{\rot{Weather}} & \multirow{5}{*}{$T_{best}$} & 24  & \textbf{0.092} & \textbf{0.129} & 0.102 & 0.146 & \underline{0.101} & 0.147 & \underline{0.102} & \underline{0.145} & 0.215 & 0.293 & 0.208 & 0.296 & 0.141 & 0.232 \\
& & 48  & \textbf{0.117} & \textbf{0.165} & 0.134 & 0.189 & \underline{0.133} & 0.188 & 0.134 & \underline{0.187} & 0.309 & 0.390 & 0.313 & 0.383 & 0.189 & 0.279 \\
& & 96  & \textbf{0.149} & \textbf{0.206} & \underline{0.169} & 0.226 & 0.170 & 0.227 & 0.170 & \underline{0.222} & 0.333 & 0.387 & 0.345 & 0.398 & 0.225 & 0.308 \\
& & 192  & \textbf{0.195} & \textbf{0.253} & 0.214 & 0.270 & \underline{0.211} & 0.264 & 0.216 & \underline{0.263} & 0.423 & 0.458 & 0.420 & 0.449 & 0.280 & 0.350 \\
& & 336  & \textbf{0.250} & \textbf{0.300} & \underline{0.257} & \underline{0.304} & 0.258 & 0.306 & 0.262 & 0.340 & 0.460 & 0.452 & 0.573 & 0.544 & 0.333 & 0.379 \\
\hline
\end{tabular}}
\caption{\textbf{Information Rich (Extensive Observation)} Multivariate time series forecasting errors in MSE and MAE, lower the better. Datasets have prediction horizon $\tau\in\{24,48,96,192,336\}$ and $T_{best}\in\{24,48,96,192,336,504\}$ represents the time window yielding the best prediction accuracy for each method and $\tau$ in these datasets. The \textbf{top results} across models are highlighted in \textbf{bold},  and second to best highlighted with \underline{underline}. Our model consistently ranks in top 2. References to SOTA methods in section 5.1.}
\end{table*}

\subsection{Spatial Route}
To improve on conventional linear layers that fail to adequately capture spatial information, we introduce a spatial route in our model. A typical linear layer maps sequence \(X_{1:T}\) to \(X_{T+1:T+\tau}\), progressing one time-step at a time for every variable in \(C\). This process neglects the intrinsic spatial information embedded within predicted sequences \(X_{T+1:T+\tau}\).\footnote{Refer to Appendix C for proof supporting this claim} To remedy this, our spatial route integrates a spatial attention (SpatialAttn) layer. This layer fine-tunes the preliminary prediction outcomes derived from an encoder-decoder series. Subsequently, a ``Residual Linear" layer is employed to generate a more refined output. This mechanism can be expressed mathematically as:
\begin{equation}
\fontsize{9pt}{10pt}\selectfont
    X^{spat}_{T+1:T+\tau} = \text{Res-L} \circ \text{SpatialAttn}(\text{Res-L}\circ\text{Res-L}(\text{PE}(X_{1:T})))\notag
\end{equation}
Notably, we also used positional embedding, the same as the one from the temporal route, here to explicitly inform our model of not only sequential information but, more importantly, channel-specific information. In the context of the aforementioned PE equations, each univariate sequence is endowed with a distinct ``mark", setting it apart from sequences corresponding to other variables. These distinct ``marks" enrich the SpatialAttn layer with auxiliary data, which becomes particularly remarkable when the layer tries to discern the influence exerted by various variables and time steps on a specific variable and its associated time instance.

\textbf{Spatial Attention: }Our SpatialAttn layer is crafted to distill the spatial dependencies, enhancing the feature representations of each spatial dimension in light of its relationship with others.

Specifically, the input to SpatialAttn, represented as \(X_{T+1:T+\tau}^{\text{prel}}\) in \(\mathbb{R}^{C \times \tau}\), undergoes a \textit{tanh} activation, scaling the values between -1 and 1. This step ensures minimal variance, thereby safeguarding numerical stability. We then delve into understanding spatial interaction. By transposing these scaled values and effecting a matrix multiplication with the originals, we derive the interaction scores: 
\begin{equation}
\fontsize{9pt}{10pt}\selectfont
\text{interact\_scores} \in \mathbb{R}^{C\times C} = \text{scores} \in \mathbb{R}^{C\times \tau} \times \text{scores}^T \in \mathbb{R}^{\tau \times C} \notag
\end{equation}
where, \begin{equation}
\fontsize{9pt}{10pt}\selectfont
    \text{scores} = tanh(X_{T+1:T+\tau}^{\text{prel}})\notag
\end{equation}
Here each element interaction\_scores$[i, j]$ represents the dot product between the temporal scores of the $i$-th spatial dimension and the $j$-th spatial dimension. The results of such matrix multiplication signify two critical features: directional similarity and influence magnitude. If two spatial dimensions often rise and fall together, their dot product will be high, indicating a high directional similarity. Furthermore, if the patterns in one variable consistently coincide with another, this matrix multiplication will capture that relationship, suggesting that one channel might be influenced by or is influencing another.

Following this, interaction scores undergo softmax normalization to produce the final attention weights, \(W \in \mathbb{R}^{C\times C}\). These weights, when multiplied with the original input and aggregated along the first dimension, indicate the collective influence of variables. Lastly, to ensure the fidelity of the SpatialAttn output, a skip connection from the input is incorporated. Mathematically, this is expressed as:
\begin{equation}
X^{attn}_{T+1:T+\tau} = X^{prel}_{T+1:T+\tau} + \sum_{i=1}^{C} W^T_{[i,:]} \times X^{i}_{T+1:T+\tau} \notag
\end{equation} 
where $W^T_{[i,:]}\in \mathbb{R}^{C\times 1}$ and $X^{i}_{T+1:T+\tau}\in\mathbb{R}^{1\times \tau}$.

\section{Experiments and Results}

\subsection{Experiment Setups}
\hspace{4.5mm}\textbf{Dataset.} we selected five well-known datasets in the field, including ETTh1 ETTm1 (electric transformer temperatures) from \cite{zhou2021informer}, Electricity and Weather, and JAAD (traffic trajectory prediction) from \cite{kotseruba2016joint}. All of them are multivariate time series, with the last dataset demonstrating a practical scenario for scarce data input. Data description can be found in Appendix A.

\textbf{Training.} To remove the error accumulation effect, we employ one-time prediction instead of incremental prediction, which is often used in transformers. As our model leverages spatiotemporal information, we deviate from the univariate training approach used by LTSF-Linear\cite{zeng2023transformers}—where the model is trained with only one channel of a sequence at a time—to an all-channel training method, in which we use the entire sequence in a single training step. Following the same work, we apply MSE loss between predicted sequence $X^{pred}_{T+1:T+\tau}$ and ground truth $X^{gt}_{T+1:T+\tau}$ to update parameters. Further implementation details and hyperparameters in our model for each dataset are elaborated in Appendix B.

\textbf{Evaluation metric.} Following previous work \cite{zeng2023transformers}, we calculate both the Mean Square Error (MSE) and the Mean Absolute Error (MAE) for all channels within each prediction sequence, which are then averaged across the entire dataset---both metrics lower the better and tests conducted on RTX 3090 GPUs.

\textbf{Compared Methods.} We compare our model with LTSF-Linear\cite{zeng2023transformers}, which includes Linear, DLinear (linear with decomposition), and NLinear (linear with normalization), and the three best transformer models (FEDformer\cite{zhou2022fedformer}, Autoformer\cite{wu2021autoformer}, Informer\cite{zhou2021informer}) from benchmarks by \cite{zeng2023transformers}.
\begin{table*}
\centering
{\fontsize{8pt}{10pt}\selectfont
\newcolumntype{C}[1]{>{\centering\arraybackslash}p{#1}}
\begin{tabular}{C{0.4cm}C{0.4cm}C{0.4cm}|*{2}{C{0.6cm}}||*{6}{C{0.5cm}}||*{6}{C{0.5cm}}}
\hline
\multicolumn{3}{c|}{Methods}  & \multicolumn{2}{c||}{\textbf{STL (Ours)}} & \multicolumn{2}{c|}{\textbf{Linear}}  & \multicolumn{2}{c|}{\textbf{DLinear}} & \multicolumn{2}{c||}{\textbf{ NLinear}} & \multicolumn{2}{c|}{\textbf{Autoformer}}  & \multicolumn{2}{c|}{\textbf{ Informer}} & \multicolumn{2}{c}{\textbf{ FEDformer}} \\
\hline
Data & $T$ &$\tau$  & MSE & MAE & MSE & MAE & MSE & MAE & MSE & MAE & MSE & MAE & MSE  & MAE & MSE & MAE\\
\hline
\multirow{5}{*}{\rot{Electricity}} & \multirow{5}{*}{48} & 24   & \textbf{0.142} & \textbf{0.237} & 0.179 & 0.261 & 0.178 & 0.260 & 0.183 & \underline{0.257} & 0.174 & 0.296 & 0.293 & 0.386 & \underline{0.158} & 0.279 \\
& & 48 & \underline{0.177} & \textbf{0.271} & 0.216 & 0.291 & 0.215 & 0.289 & 0.224 & 0.287 & 0.199 & 0.313 & 0.306 & 0.396 & \textbf{0.175} & \underline{0.291} \\
& & 96 & \textbf{0.192} & \textbf{0.282} & 0.246 & 0.316 & 0.246 & 0.315 & 0.258 & 0.312 & 0.225 & 0.331 & 0.323 & 0.408 & \textbf{0.192} & \underline{0.307} \\
& & 192 & \textbf{0.192} & \textbf{0.283} & 0.231 & 0.309 & 0.230 & 0.308 & 0.241 & \underline{0.306} & 0.244 & 0.350 & 0.360 & 0.431 & \underline{0.203} & 0.316 \\
& & 336 & \textbf{0.206} & \textbf{0.297} & 0.247 & 0.326 & 0.247 & 0.325 & 0.261 & \underline{0.323} & 0.257 & 0.357 & 0.372 & 0.441 & \underline{0.217} & 0.329 \\
\hline
\multirow{5}{*}{\rot{Etth1}} & \multirow{5}{*}{48} & 24 & \textbf{0.310} & \textbf{0.353} & 0.329 & 0.365 & 0.318 & 0.356 & 0.325 & \underline{0.360} & 0.407 & 0.444 & 0.533 & 0.530 & \underline{0.314} & 0.376 \\
& & 48 & \underline{0.347} & \textbf{0.372} & 0.363 & 0.386 & 0.352 & 0.376 & 0.361 & \underline{0.381} & 0.454 & 0.471 & 0.625 & 0.549 & \textbf{0.343} & 0.391\\
& & 96 & \underline{0.391} & \textbf{0.401} & 0.405 & 0.411 & 0.396 & 0.402 & 0.405 & \underline{0.406} & 0.462 & 0.469 & 0.752 & 0.611 & \textbf{0.384} & 0.417 \\
& & 192 & \underline{0.444} & \textbf{0.437} & 0.455 & 0.443 & 0.448 & 0.437 & 0.458 & \underline{0.438} & 0.537 & 0.520 & 1.004 & 0.748 & \textbf{0.429} & 0.449 \\
& & 336 & \underline{0.499} & \underline{0.473} & 0.507 & 0.476 & 0.500 & 0.469 & 0.509 & \textbf{0.466} & 0.524 & 0.514 & 1.108 & 0.810 & \textbf{0.476} & 0.475 \\
\hline
\multirow{5}{*}{\rot{Ettm1}} & \multirow{5}{*}{48} & 24 & \underline{0.403} & \textbf{0.394} & 0.543 & 0.453 & 0.539 & 0.451 & 0.581 & 0.452 & \textbf{0.402} & 0.433 & 0.460 & 0.439 & 0.405 & \underline{0.404} \\
& & 48 & \underline{0.485} & \underline{0.440} & 0.623 & 0.503 & 0.617 & 0.500 & 0.670 & 0.504 & 0.518 & 0.495 & 0.657 & 0.537 & \textbf{0.451} & \textbf{0.430} \\
& & 96 & \underline{0.413} & \textbf{0.407} & 0.502 & 0.449 & 0.499 & 0.449 & 0.532 & 0.450 & 0.647 & 0.544 & 0.671 & 0.572 & \textbf{0.401} & \underline{0.415} \\
& & 192 & \textbf{0.439} & \textbf{0.431} & 0.539 & 0.470 & 0.537 & 0.469 & 0.577 & 0.473 & 0.757 & 0.585 & 0.830 & 0.620 & \underline{0.462} & \underline{0.455} \\
& & 336 & \textbf{0.458} & \textbf{0.449} & 0.583 & 0.498 & 0.580 & 0.495 & 0.629 & 0.503 & 0.571 & 0.540 & 1.003 & 0.733 & \underline{0.495} & \underline{0.478} \\
\hline
\multirow{5}{*}{\rot{Weather}} & \multirow{5}{*}{48} & 24 & \textbf{0.115} & \underline{0.165} & 0.131 & 0.193 & 0.124 & 0.178 & \underline{0.121} & \textbf{0.142} & 0.315 & 0.389 & 0.315 & 0.387 & 0.141 & 0.232 \\
& & 48 & \textbf{0.160} & \underline{0.216} & 0.187 & 0.260 & 0.182 & 0.257 & \underline{0.170} & \textbf{0.199} & 0.571 & 0.535 & 0.526 & 0.526 & 0.189 & 0.279 \\
& & 96 & \textbf{0.196} & \underline{0.263} & 0.233 & 0.308 & 0.225 & 0.293 & \underline{0.215} & \textbf{0.246} & 0.504 & 0.483 & 0.454 & 0.487 & 0.231 & 0.312 \\
& & 192 & \textbf{0.228} & \underline{0.290} & 0.265 & 0.331 & 0.256 & 0.320 & \underline{0.258} & \textbf{0.280} & 0.691 & 0.560 & 0.574 & 0.548 & 0.483 & 0.478 \\
& & 336 & \textbf{0.275} & \underline{0.327} & \underline{0.313} & 0.370 & 0.307 & 0.361 & 0.315 & \textbf{0.323} & 0.679 & 0.593 & 0.609 & 0.568 & 0.357 & 0.390 \\
\hline

\multirow{3}{*}{\rot{JAAD$^*$}} & 10 & 45 & \textbf{0.277} & \textbf{8.81} & 1.40 & 17.5 & 0.620 & 12.2 & 0.500 & 10.9 & 0.845 & 16.3 & 0.296 & 9.90 & \underline{0.280} & \underline{8.90} \\
& 15 & 45 & \underline{0.300} &\textbf{ 8.75} & 1.50 & 18.5 & 0.690 & 13.0 & 0.391 & 9.60 & 0.817 & 16.4 & \textbf{0.286} & \underline{9.50 }&  \underline{0.300} & 9.70\\
& 30 & 45 & \textbf{0.371} & \underline{10.1} & 2.20 & 22.3 & 1.29 & 17.4 & 0.389 & \textbf{9.40} & 1.70 & 21.1 & \underline{0.377} & 12.2 & 0.567 & 12.9 \\
\hline
\end{tabular}}
\caption{\textbf{Information Scarce (Fixed Limited Observation)} Multivariate time series forecasting errors in MSE and MAE. Among them, JAAD dataset* is tested with an observation horizon $T\in \{10, 15, 30\}$ for prediction horizon $\tau=45$ (For JAAD dataset, the values are provided in units of $10^{-3}$). Other datasets have the same observation horizon $T=48$ and prediction horizons $\tau\in\{24,48,96,192,336\}$. The \textbf{top results} are highlighted in \textbf{bold}, and second to best highlighted with \underline{underline}. Our model consistently ranks in top 2. References to SOTA methods in section 5.1.}
\end{table*}
\subsection{Information-Rich Scenarios}
\hspace{5mm}\textbf{Setup.} We follow \cite{zeng2023transformers} for testing information-rich scenarios: models report the best prediction accuracy of a prediction horizon $\tau$ from a variety of observation lengths $T\in\{24,48,96,192,336,504\}$ on the first four datasets. To further understand a model's adaptability for both short and long-term forecasting, we expanded the prediction horizon to encompass lengths $\tau\in\{24,48,96,192,336\}$.

\textbf{The results}, shown in Table.1, demonstrate our model's ability to surpass other models in data-rich prediction tasks, irrespective of the forecasting duration. Noteworthy from the table, LTSF-Linear models indeed surpass the performance of transformer-based models by quite a margin. Our STL further extends to such superiority, providing an up to 14.3\% boost in MSE to the top-performing DLinear.
\subsection{Information Scarce Experiments}
\hspace{4.5mm}\textbf{Setup.} To prove the robustness of our model in situations characterized by limited data availability, we simulate data-scarce scenarios: we fix the observation horizon at $T=48$ and evaluate prediction accuracy over a range of prediction lengths, $\tau\in\{24,36,48,72,96,120,144,168,192,336\}$, using the first four datasets. Depending on the dataset's recording interval, This observation horizon translates to either a 2-day or a 6-hour observation window. Such brief observational periods are not rare in real-world scenarios. For instance, if a new infectious disease arises, medical professionals may have to evaluate the severity based on just a few days of data. Similarly, when new stocks hit the market, financial analysts might base their investment decisions on the stock's debut day performance.

Broadening the scope, we also evaluated our model's performance on the JAAD dataset, in alignment with the approach delineated in \cite{rasouli2019pie}. We set a prediction horizon of 45 time steps, equivalent to 1.5 seconds. We then trialed various observation lengths lengths, $T\in\{30(1s),15(0.5s),10(0.33s)\}$, to further underscore the model's applicability in real-world, data-scarce situations.

\textbf{Noticeable degradation} in prediction accuracy from the former benchmark: DLinear lags considerably behind transformer-based alternatives with an average \textbf{48\% deterioration} in MSE under these conditions. Contrastingly, our model demonstrates remarkable resilience, delivering up to a \textbf{34\% enhancement} in MSE to DLinear across the four datasets used in \cite{zeng2023transformers}. For real-world, data-limited settings of JAAD, STL yields a more impressive 55\% uptick in predictive accuracy over DLinear. Such a huge difference stresses the need to pay attention to observation length when designing models for time series forecasting tasks.

\textbf{Visualize.} To illustrate STL's resilience to data scarcity, we plot prediction accuracy (in $\frac{1}{\text{MSE}}$) for all models with a) observation length $T=48$ and varying prediction lengths $\tau\in\{24,36,48,72,96,120,144,168,192,336\}$, and b) fixed prediction length $\tau=336$ against diverse observation length $T\in\{24,36,48,72,96,120,144,168,336,504\}$ in Fig.1, utilizing the Weather dataset in both scenarios. Here, a higher $\frac{1}{\text{MSE}}$ implies a reduced MSE and a prediction that more closely aligns with the ground truth. From the visual data, STL consistently outperforms other models under an observation window of $T=48$ (Fig.1(a)). Moreover, the drop in STL's prediction accuracy is more gradual than that of LTSF Linear models when the observation length decreases from 504 to 24 (Fig.1(b)).

\subsection{Ablation Study}
\hspace{4.5mm}\textbf{Setup.} To delve into the contribution of each design component in our model, we conducted an ablation study. Specifically, we assessed the importance of the three routes by successively excluding them from the STL. This resulted in four distinct configurations: [Full STL; Core + Spatial (with Temporal removed); Core-only; Basic Linear layer].

Our experiments spanned prediction lengths of $\tau\in\{24,36,48,72,96,120,144,168,192,336\}$ with an observation length of $T=48$ on two datasets: Electricity and Weather. Results are shown in Fig.4, which depicts the prediction sequence $\tau$ plotted against $\frac{1}{\text{MSE}}$. A higher value of $\frac{1}{\text{MSE}}$ signifies better prediction accuracy.

\textbf{Results.} In Fig.4, across the various prediction horizons, there emerges a discernible trend: as modules are reintegrated (progressing from the basic linear layer in green to full STL in purple), prediction accuracy exhibits a \textbf{sequential increase}. Compared to the baseline Linear model, this substantial uplift in performance is consistent with our theoretical perspective: A linear layer only conducts regression on partial spatiotemporal information, represented as $\beta(s',t')$. By considering the omitted information via the spatial and temporal routes, STL attains a more sophisticated and robust regression, $\beta(s,t)$.
\begin{figure}[h]
    \centering
    \includegraphics[width=1\columnwidth]{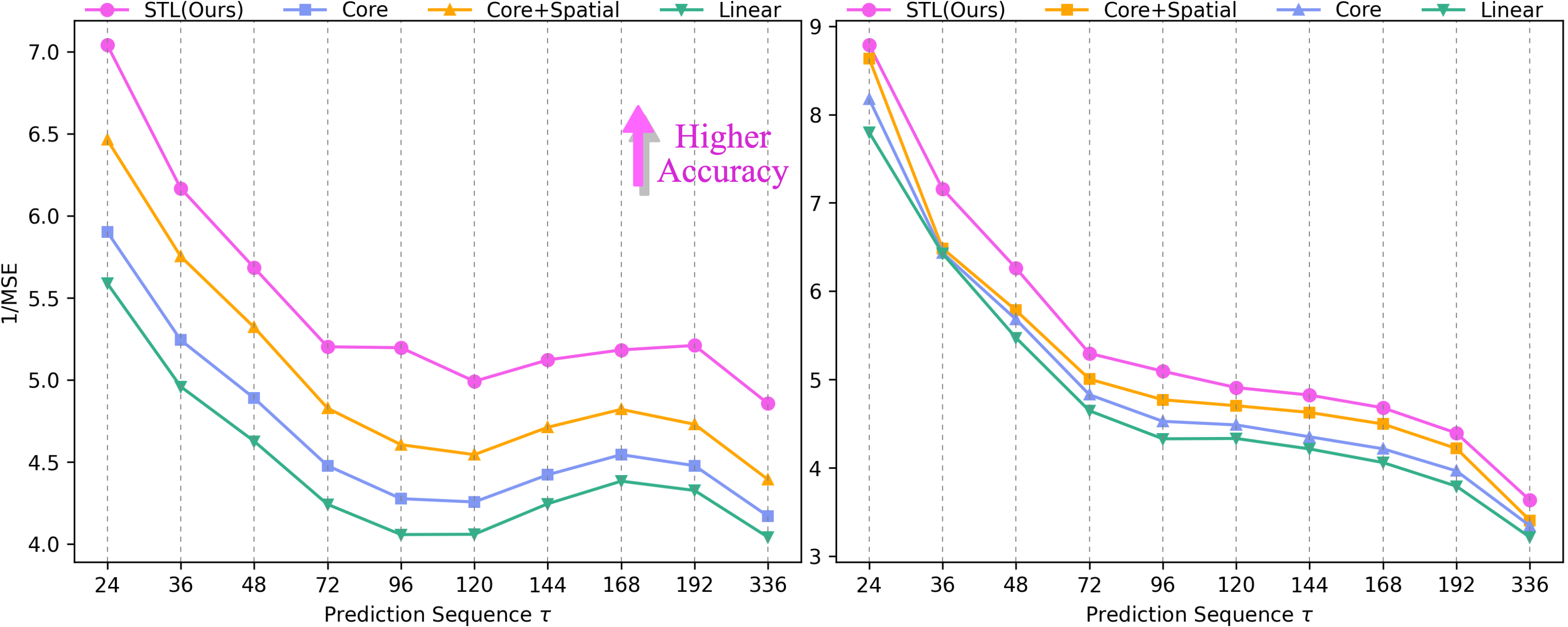}
    \caption{Plot of STL's ablation study on Electricity (Left) and Weather (Right) Dataset with observation $T=48$ and prediction $\tau\in\{24, 36, 48, 72, 96, 120, 144, 168, 192, 336\}$. Zoom in for detail.}
    \label{fig:enter-label}
\end{figure}
\begin{figure}[h]
    \centering
    \includegraphics[width=1\columnwidth]{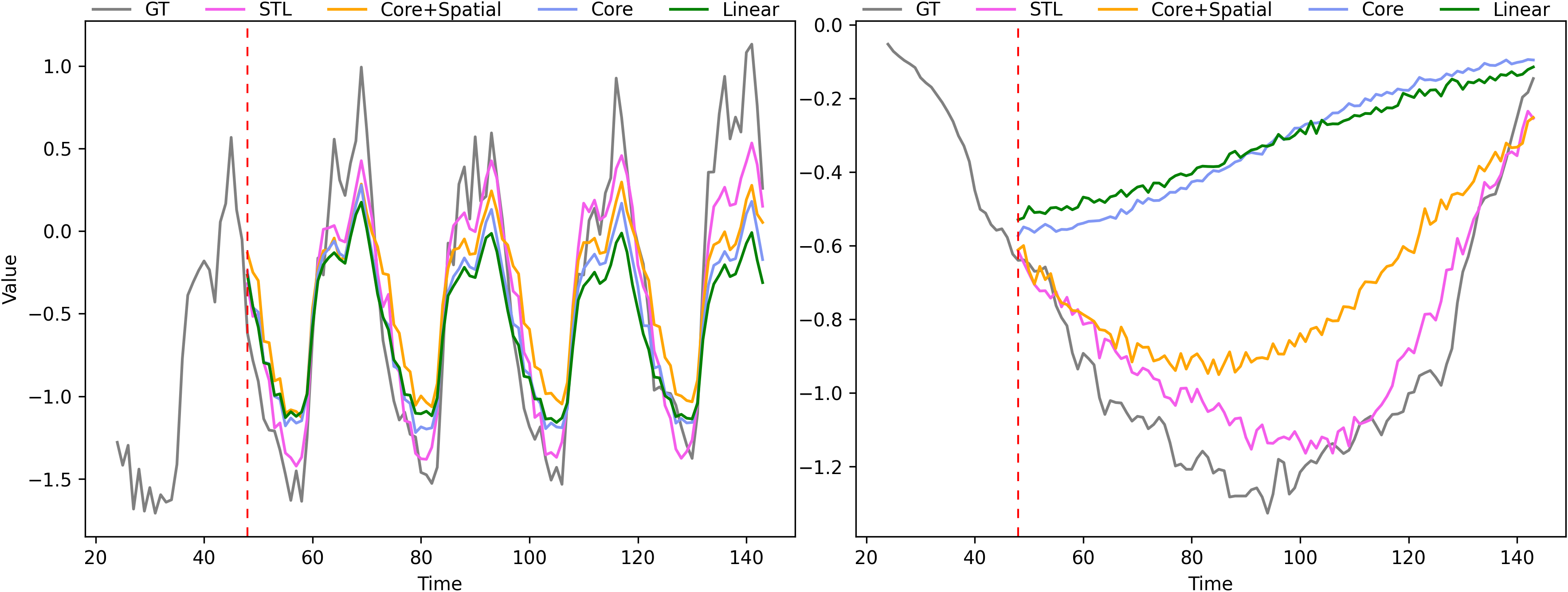}
    \caption{Visualization of STL's ablation study on Electricity (Left) and Weather (Right) Dataset with observation $T=48$ and prediction $\tau=192$ (only first 96 prediction shown due to space). Zoom in for detail.}
    \label{fig:enter-label}
\end{figure}

\textbf{Visualize.} To further underscore the significance of each route, Fig.5 provides a visual comparison of prediction outcomes for two select cases for Electricity and Weather datasets. Here, prediction horizon $\tau=192$ with the prediction of one variable is shown (More visualization in Appendix A). The observed pattern is consistent with our numerical findings: as routes are reincorporated, predictions progressively approximate the ground truth (in grey) more closely. Specifically, the left pane of Fig.5 underscores the ``refinement" capability introduced by skip connections and the effectiveness of the ``coarse-refine" strategy in the spatial route. Additionally, the right pane emphasizes the rectification provided by the spatial and temporal routes: when the relationship captured by the Core route is skewed, these auxiliary routes help rectify inappropriate predictions while implicitly guiding the Core route for other dependencies.

\section{Conclusion}
\hspace{4.5mm}\textbf{Model.} In the work, we propose the SpatioTemporal-Linear (STL) framework that leverages temporal and spatial dynamics to enhance linear-based prediction. Notably, STL not only presents an innovative method for embedding spatiotemporal information but also extends the superiority of linear-based models over more sophisticated architectures like transformers. Our extensive evaluations validate STL's adaptability across varied prediction durations and its robustness to diverse data richness. 

\textbf{Impacts.} Through this discourse, we highlight two crucial insights: First, integrating both temporal and spatial information into forecasting models holds substantial value, meriting as much attention as the intrinsic data points of a series. Second, we emphasize the need to study the influence of data scarcity on time series forecasting models' prediction results for truly universal applications across domains.

\bibliographystyle{unsrt}
\bibliography{references}

\section{Appendix A Data Description}
We follow \cite{zeng2023transformers} to derive the following datasets: ETTh1 and ETTm1 (Electricity transformer temperature)\footnote{https://github.com/zhouhaoyi/ETDataset}, Electricity\footnote{\url{https://archive.ics.uci.edu/ml/datasets/ElectricityLoadDiagrams20112014}}, and Weather\footnote{https://www.bgc-jena.mpg.de/wetter/}. JAAD dataset, on the other hand, has ego-centric videos captured at 30 frames per second, in which pedestrians' locations are labeled by bounding boxes $[x_1,y_1,x_2,y_2]$. As each pedestrian is unique, we consider how many trajectories of 60 time-steps are available following \cite{rasouli2019pie}. The features of these datasets are listed in the table below:

\begin{table}[h]
    \centering
    {\fontsize{9pt}{10pt}\selectfont
    \newcolumntype{C}[1]{>{\centering\arraybackslash}p{#1}}
    \begin{tabular}{C{1.2cm}|C{1.1cm}|C{1.5cm}|C{1.4cm}|C{1.0cm}}
    \toprule
        \multicolumn{1}{C{1.2cm}|}{Dataset} & \multicolumn{1}{C{1.1cm}|}{Variables} & \multicolumn{1}{C{1.5cm}|}{Time Steps} & \multicolumn{1}{C{1.4cm}|}{Interval} & \multicolumn{1}{C{1.0cm}}{Ratio*} \\
    \midrule
         Electricity & 321 & 26304 & 1 Hr. & 6:2:2\\ 
         ETTh1 & 7 & 17420 & 1 Hr. & 6:2:2\\
         ETTm1 & 7 & 69680 & 15 Min. & 7:1:2\\
         Weather & 21 & 52696 & 1 Hr. & 5:1:4\\
         JAAD & 4 & 2800 (traj.) & 0.033 Sec. & 7:1:2\\
    \bottomrule
    \end{tabular}}
    \caption{Summary of Datasets' Features. Ratio* represents the $Train:Validation:Test$}
    \label{Dataset Features}
\end{table}

\section{Appendix B Implementation Details}
\subsection{Batched Dataloader}
Throughout the paper, our model's input and output are presented in a single-instance format, equivalent to a batch size of 1. For more efficient training during implementation, we utilize a batched approach. In this context, each batch contains $B$ multivariate time series instances, represented as $X_{1:T+\tau} \in \mathbb{R}^{\tau\times C}$. Thus, the model's observation input becomes $[B, X_{1:T} \in \mathbb{R}^{T\times C}]$, and its prediction output is $[B, X_{T+1:T+\tau} \in \mathbb{R}^{\tau\times C}]$. The batch size $B$ is a hyperparameter, consistently set to 32 for all training and testing phases of our model. For reference models, we adopt the configurations from \cite{zeng2023transformers}.
In terms of time embeddings, they are batched similarly as $[B, \text{date-time}]$. The exact shape of these embeddings varies with the dataset in use. Specifically, the ETTh1, Electricity, and Weather datasets incorporate [month; day; weekday; hour] date-time stamps, while ETTm1 adds an additional [minute] stamp. Notably, the JAAD dataset does not feature any date-time stamps.
\subsection{Hyperparameters}
We set the torch random seed of all models to 2021 following \cite{zeng2023transformers}. Table.4 delineates hyperparameters' values in our model that vary on different datasets.
\begin{table}[h]
    \centering
    {\fontsize{9pt}{10pt}\selectfont
    \newcolumntype{C}[1]{>{\centering\arraybackslash}p{#1}}
    \begin{tabular}{C{1.1cm}|C{1.5cm}|C{0.8cm}|C{0.7cm}|C{0.7cm}|C{1.1cm}}
    \toprule
        \multicolumn{1}{C{1.1cm}|}{Dataset} & \multicolumn{1}{C{1.5cm}|}{$hidden\_size$} & \multicolumn{1}{C{0.8cm}|}{dropout} & \multicolumn{1}{C{0.7cm}|}{LR} & \multicolumn{1}{C{0.7cm}|}{decay} &\multicolumn{1}{C{1.1cm}}{act.}\\
    \midrule
         Electricity & 512 & 0.0 & 6e-4 & 0.8&SiLU \\
         ETTh1 & 256 & 0.1 & 2e-4 & 0.75 &LeakyReLU\\
         ETTm1 & 256 & 0.25 & 2e-4 & 0.8 &SiLU\\
         Weather & 512 & 0.25 & 2e-4 & 0.75 &SiLU\\
         JAAD & 512 & 0.0 & 1e-3 & 0.9 &SiLU\\
    \bottomrule
    \end{tabular}}
    \caption{Hyperparameters of Best Configuration for Each Dataset.}
    \label{Hyperparameters of Best Configuration for Each Dataset}
\end{table}

\section{Appendix C Proof of Fragmentary Regression Coefficients}
Our aim is to demonstrate that regression coefficients derived from simple univariate linear layers are not exhaustive in capturing comprehensive spatiotemporal dynamics. Given that linear regression is a canonical method, our focus shifts to establishing that such layers fail to encapsulate the entirety of the spatiotemporal information, denoted as \( s' \subset s, t' \subset t \).

Initiating our proof, consider a standard regression equation used by linear layers. A linear layer predicts a future point \( x_{T+1} \) based on a univariate observation sequence \( x_{1:T} \) as:
\begin{equation}
    x_{T+1} = \sum_{i=1}^{T} x_i \cdot w_i + b
\end{equation}
\begin{figure*}[h!]
    \centering
    \includegraphics[width=1\columnwidth]{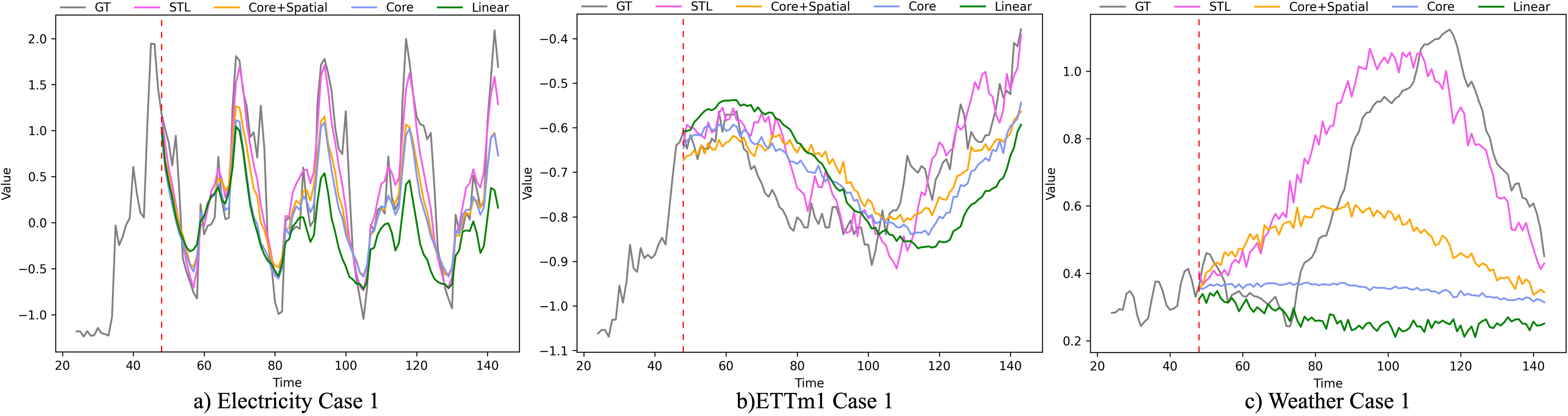}
    \label{fig:enter-label}
    \includegraphics[width=1\columnwidth]{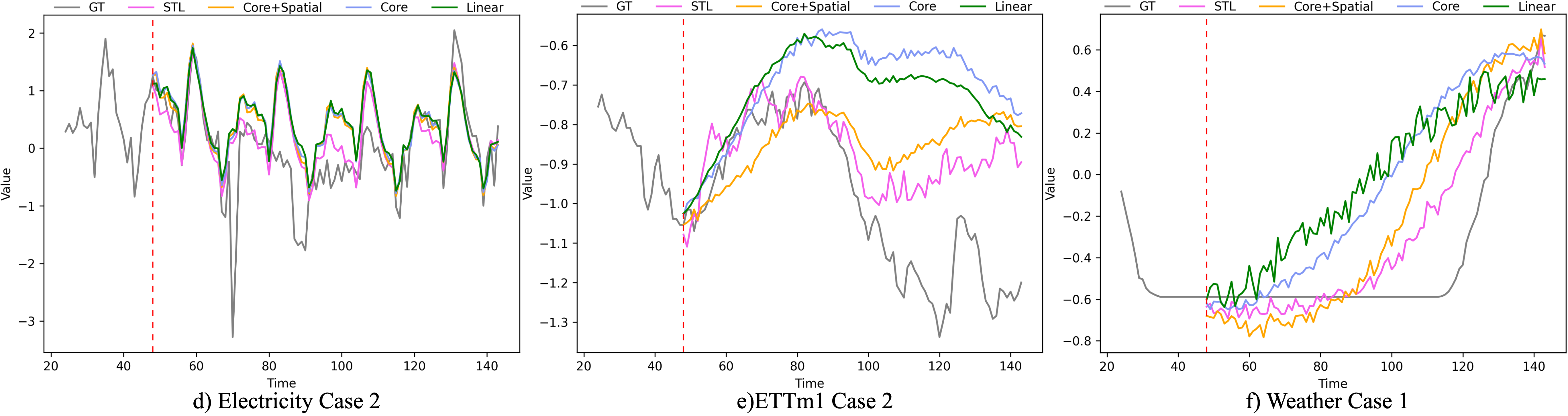}
    \label{fig:enter-label}
    \includegraphics[width=1\columnwidth]{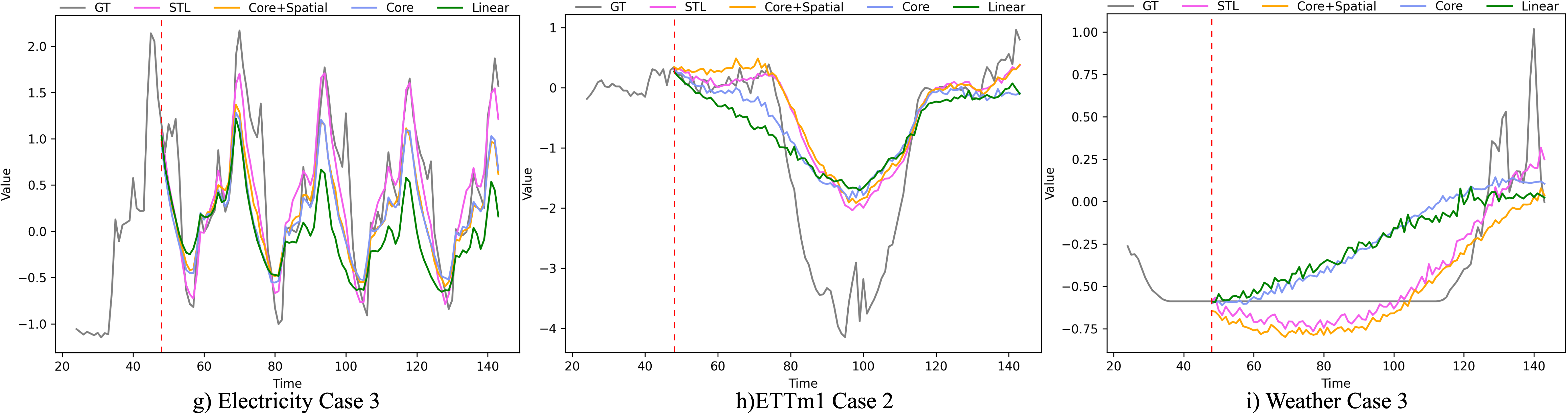}
    \label{fig:enter-label}
    \includegraphics[width=1\columnwidth]{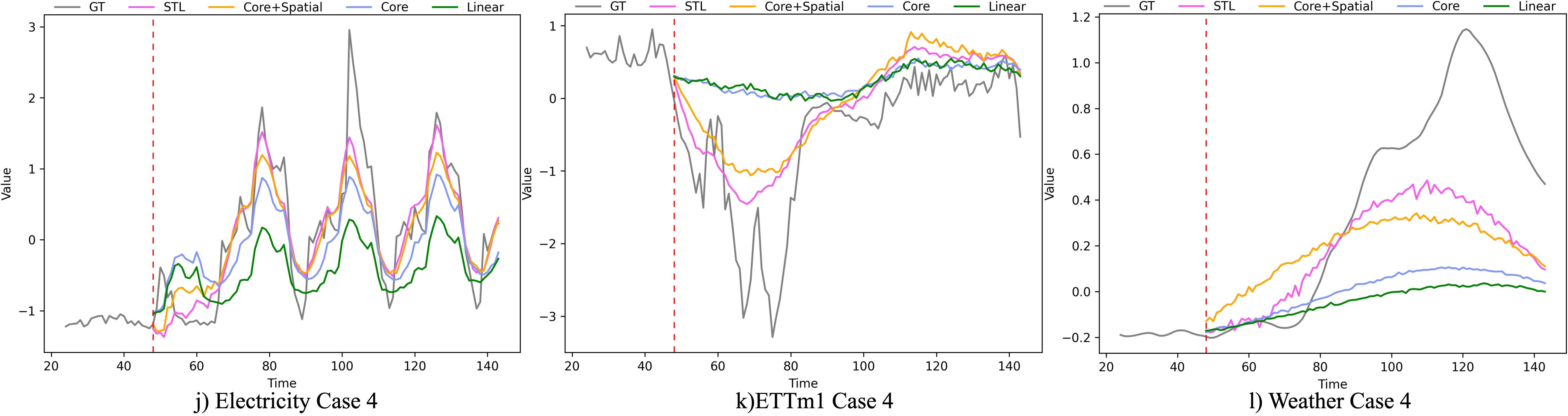}
    \caption{Visualization of STL’s ablation study on Electricity
    (Left), Etthm1 (Middle), and Weather (Right) Dataset with observation length $T=48$ and prediction length $\tau=192$ on different variables and starting time.}
    \label{fig:AppendixD}
\end{figure*}
While a linear layer does assign varying weights to different elements in the sequence, encapsulating some temporal nuances, it predominantly regresses based on the magnitudes of \( x_i \) rather than their positions. Besides, such layers neglect inherent chronological properties---knowing, for instance, that \( x_T \) represents electric utility data logged at 4 P.M. on a Friday should prompt an anticipation of increased usage in subsequent hours. Thus, the layer captures merely a subset of the temporal information embedded in the dataset $t'|t' \subset t$.

Considering spatial dimensions, linear layers also exhibit deficiencies. Given a multivariate observation sequence with \( C \) variables, described as \( X_{1:T} \in \mathbb{R}^{T\times C} \), each element \( X_{\delta c} \) is contingent on past values across all channels:
\begin{equation}
    X_{\delta c} = f\left( \bigcup_{j=1}^{\delta} \bigcup_{d=1}^{C} X_{jd} \right)|\delta \leq T
\end{equation}
As an illustrative example, Company A's stock price at 2 P.M. could be influenced by the morning's trading patterns of correlated stocks, exchange rate shifts, and commodity valuations. Linear layers, when using \( x_{1:T} \) to predict \( x_{T+1} \), do capture certain spatial dynamics that have historically influenced values. However, their predictive capacity does not extend to future interdependencies, such as the influence of proximal future points on distal ones:
\begin{equation}
    X_{\delta c} = f\left( \bigcup_{j=T+1}^{\delta} \bigcup_{d=1}^{C} X_{jd} \right)|T+1\leq\delta\leq T+\tau
\end{equation}

Losing this information, the spatial information that a linear layer captured is only $s'|s' \subset s$. 
\begin{figure*}[t]
    \centering
    \includegraphics[width=1\columnwidth]{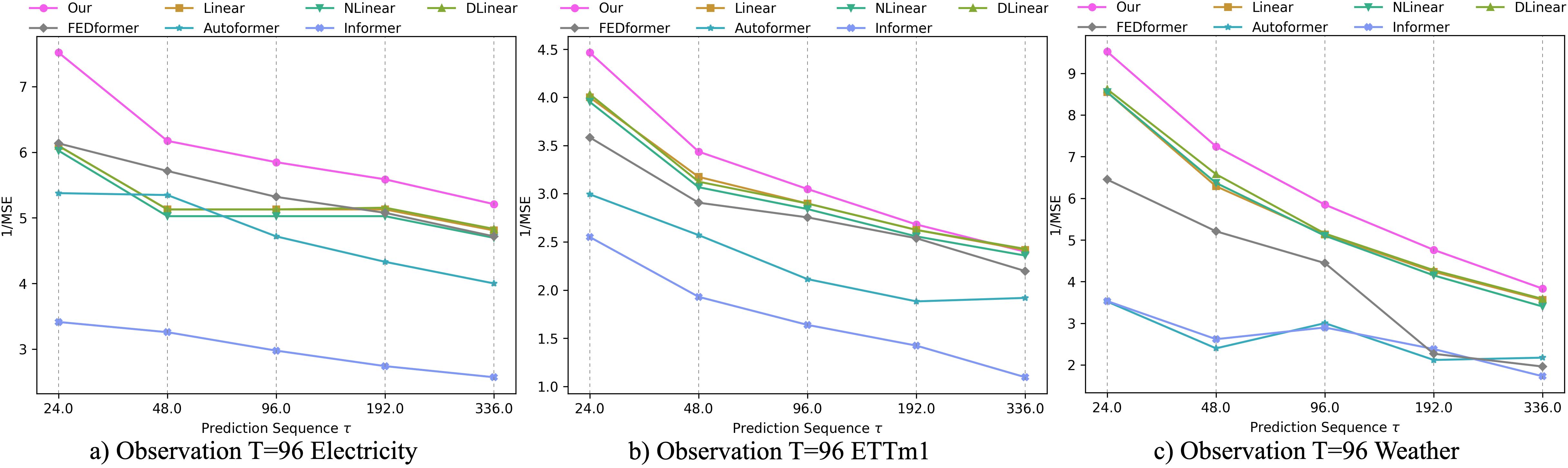}
    \includegraphics[width=1\columnwidth]{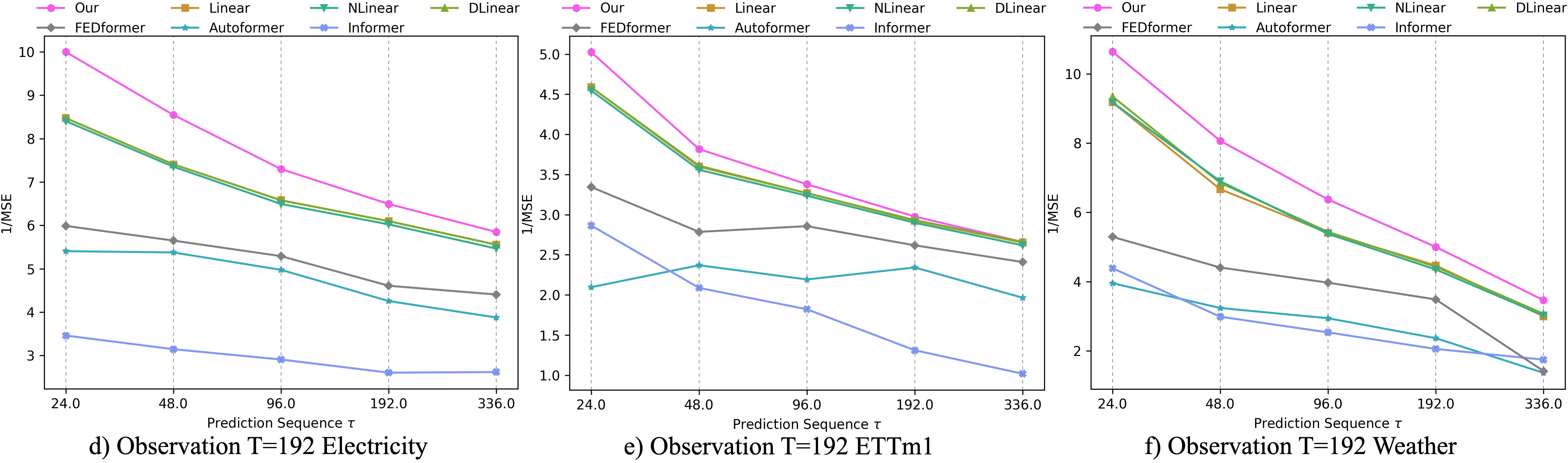}
    \includegraphics[width=1\columnwidth]{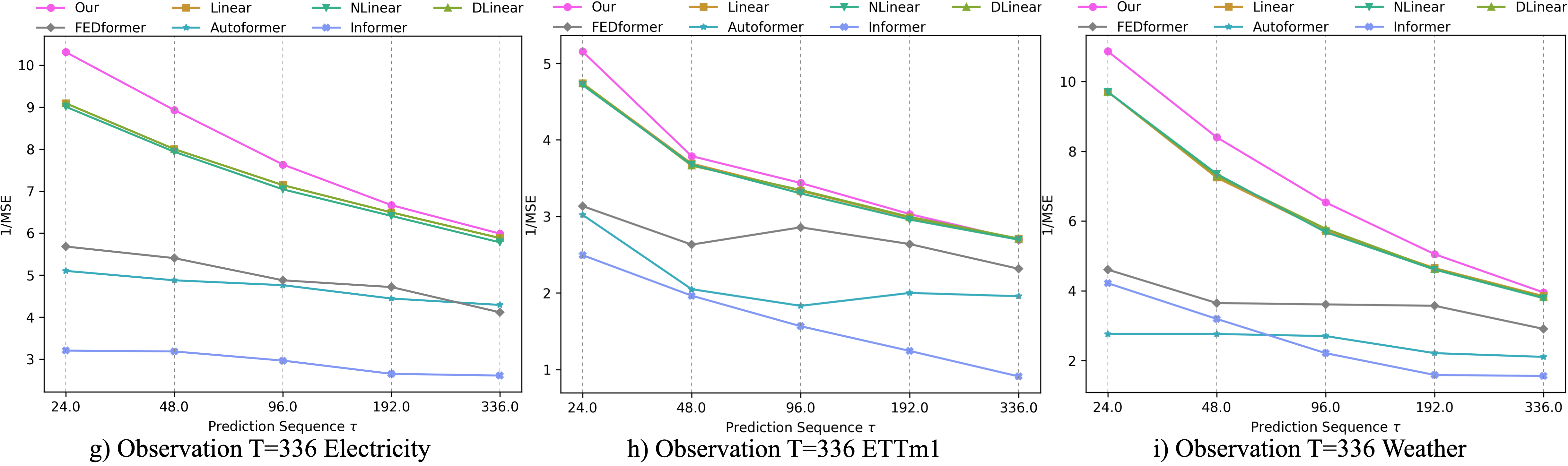}
    \caption{Plots of STL’s ablation study on Electricity (Left), Ettm1 (Middle), and Weather (Right) Dataset with observation $T\in\{96, 192, 336\}$ and
    $\tau\in\{24,48,96, 192, 336\}$.}
    \label{fig:AppendixE}
\end{figure*}
Thus far, we have proved that a simple linear layer only regress incomplete spatiotemporal information from $X_{1:T} \in \mathbb{R}^{T\times C}$ to attain future sequences $X_{T+1:T+\tau} \in \mathbb{R}^{\tau\times C}$, and thus the coefficients derived can be denoted as $\beta(s', t')|s' \subset s, t' \subset t$.

\section{Appendix D Extra Ablation Study Visualization}
We showcase an additional four prediction cases with sequence length $T=48$ and prediction length $\tau=192$ in each dataset of Electricity, ETTm1, and Weather in Fig.6. Gray lines represent ground truth, and STL in purple clearly approximates the ground truth with the highest precision.

\section{Appendix E Extra Experimental Visualization on Various Observation and Prediction}
We showcase, in Fig.7,  an additional three visualization with $1/\text{MSE}$ against prediction length $\tau\in\{24,48,96,192,336\}$ under various observation horizons $T\in\{96,192,336\}$ on Electricity, ETTm1, and Weather dataset. Under all these conditions, STL achieves supreme prediction results.

\section{Appendix F Extra Numerical Experimental Results on Various Observation and Prediction}
We display the numerical prediction results measured in MSE and MAE for varied prediction lengths $\tau\in\{24,48,96,192,336\}$ under observation lengths $T\in\{24,96,192,336,504\}$ from Table.5-9.
\begin{table*}[h!]
\centering
{\fontsize{9pt}{10pt}\selectfont
\newcolumntype{C}[1]{>{\centering\arraybackslash}p{#1}}
\begin{tabular}{C{0.4cm}C{0.5cm}C{0.5cm}|*{2}{C{0.7cm}}||*{6}{C{0.6cm}}||*{6}{C{0.6cm}}}
\hline
\multicolumn{3}{c|}{Methods}  & \multicolumn{2}{c||}{\textbf{STL (Ours)}} & \multicolumn{2}{c|}{\textbf{Linear}}  & \multicolumn{2}{c|}{\textbf{DLinear}} & \multicolumn{2}{c||}{\textbf{ NLinear}} & \multicolumn{2}{c|}{\textbf{Autoformer}}  & \multicolumn{2}{c|}{\textbf{ Informer}} & \multicolumn{2}{c}{\textbf{ FEDformer}} \\
\hline
Data & $T$ &$\tau$  & MSE & MAE & MSE & MAE & MSE & MAE & MSE & MAE & MSE & MAE & MSE  & MAE & MSE & MAE\\
\hline
\multirow{5}{*}{\rot{Electricity}} & \multirow{5}{*}{24}  & 24 & \underline{0.175} & \textbf{0.266} & 0.200 & 0.278 & 0.198 & 0.276 & 0.207 & \underline{0.273} & 0.185 & 0.304 & 0.284 & 0.375 & \textbf{0.163} & 0.282 \\
& & 48 & \underline{0.202} & \textbf{0.290} & 0.243 & 0.312 & 0.241 & 0.310 & 0.256 & 0.306 & 0.289 & 0.386 & 0.297 & 0.386 & \textbf{0.186} & \underline{0.299} \\
& & 96 & \underline{0.220} & \textbf{0.310} & 0.277 & 0.339 & 0.276 & 0.337 & 0.298 & 0.335 & 0.284 & 0.378 & 0.311 & 0.396 & \textbf{0.200} & \underline{0.311} \\
& & 192 & \underline{0.218} & \textbf{0.304} & 0.268 & 0.338 & 0.267 & 0.336 & 0.286 & 0.333 & 0.275 & 0.367 & 0.392 & 0.453 & \textbf{0.207} & \underline{0.318} \\
& & 336 & \underline{0.231} & \textbf{0.317} & 0.286 & 0.355 & 0.285 & 0.353 & 0.309 & 0.352 & 0.288 & 0.378 & 0.403 & 0.461 &\textbf{0.221} & \underline{0.331}\\
\hline

\multirow{5}{*}{\rot{Etth1}} & \multirow{5}{*}{24} & 24 & \underline{0.353} & \textbf{0.383}& 0.377 & 0.396 & 0.363 & \underline{0.386} & 0.376 & 0.391 & 0.392 & 0.424 & 0.518 & 0.512 &\textbf{0.330}& 0.387 \\
& & 48 & \underline{0.391} & \underline{0.401} & 0.410 & 0.412 & 0.397 & 0.402 & 0.412 & 0.408 & 0.472 & 0.471 & 0.629 & 0.569 & \textbf{0.361} & \textbf{0.399} \\
& & 96 & \underline{0.429} & \textbf{0.427} & 0.450 & 0.438 & 0.436 & \underline{0.428} & 0.454 & 0.433 & 0.565 & 0.511 & 0.799 & 0.653 & \textbf{0.399} & \underline{0.428} \\
& & 192 & \underline{0.480} & 0.464 & 0.501 & 0.471 & 0.488 & \underline{0.462} & 0.508 & 0.465 & 0.571 & 0.543 & 0.979 & 0.748 &\textbf{0.446}& \textbf{0.454} \\
& & 336 & \underline{0.529} & 0.496 & 0.551 & 0.503 & 0.539 & 0.495 & 0.560 & \underline{0.494} & 0.637 & 0.562 & 1.106 & 0.802 & \textbf{0.492} &\textbf{0.454} \\
\hline
\multirow{5}{*}{\rot{Ettm1}} & \multirow{5}{*}{24}& 24 & \underline{0.403} & \underline{0.410} & 0.555 & 0.459 & 0.552 & 0.456 & 0.631 & 0.461 & 0.409 & 0.433 & 0.463 & 0.445 & \textbf{0.398} & \textbf{0.386} \\
& & 48 & \textbf{0.530} & \underline{0.478} & 0.799 & 0.582 & 0.791 & 0.581 & 0.957 & 0.598 & 0.621 & 0.563 & 0.698 & 0.566 & \textbf{0.561} & \textbf{0.472} \\
& & 96 & \textbf{0.482} & \textbf{0.450} & 0.671 & 0.532 & 0.675 & 0.529 & 0.787 & 0.541 & 0.743 & 0.587 & 0.760 & 0.597 & \underline{0.504} & \underline{0.456} \\
& & 192 & \textbf{0.512} & \textbf{0.467} & 0.704 & 0.552 & 0.701 & 0.549 & 0.833 & 0.567 & 0.755 & 0.597 & 0.899 & 0.675 & \underline{0.540} & \underline{0.486} \\
& & 336 &\textbf{0.528} & \textbf{0.492} & 0.742 & 0.576 & 0.738 & 0.574 & 0.886 & 0.597 & 0.830 & 0.629 & 1.024 & 0.736 & \underline{0.586} & \underline{0.511} \\
\hline
\multirow{5}{*}{\rot{Weather}} & \multirow{5}{*}{24} & 24 & \textbf{0.115} & \underline{0.170} & 0.128 & 0.187 & 0.126 & 0.186 & \underline{0.123} & \textbf{0.139} & 0.215 & 0.293 & 0.277 & 0.354 & 0.146 & 0.214 \\
& & 48 & \underline{0.181} & \underline{0.249} & 0.196 & 0.271 & 0.189 & 0.260 & \textbf{0.172} & \textbf{0.196} & 0.677 & 0.549 & 0.483 & 0.508 & 0.232 & 0.304 \\
& & 96 & \textbf{0.225} & \underline{0.298} & 0.259 & 0.335 & 0.258 & 0.333 & \textbf{0.226} & \textbf{0.253} & 0.534 & 0.511 & 0.647 & 0.589 & 0.304 & 0.362 \\
& & 192 & \textbf{0.246} & \underline{0.310} & 0.281 & 0.348 & 0.281 & 0.349 & \underline{0.270} & \textbf{0.286} & 0.592 & 0.547 & 0.580 & 0.543 & 0.339 & 0.376 \\
& & 336 & \textbf{0.289} & \underline{0.340} & 0.334 & 0.390 & 0.326 & 0.381 & \underline{0.331} & \textbf{0.329} & 0.734 & 0.595 & 0.573 & 0.544 & 0.707 & 0.574 \\
\hline
\end{tabular}}
\caption{Multivariate time series forecasting errors in MSE and MAE. Observation horizon $T=24$ and prediction horizons $\tau\in\{24,48,96,192,336\}$.The \textbf{top results} are highlighted in \textbf{bold}, and second to best highlighted with \underline{underline}.}
\end{table*}
\begin{table*}[h!]
\centering
{\fontsize{9pt}{10pt}\selectfont
\newcolumntype{C}[1]{>{\centering\arraybackslash}p{#1}}
\begin{tabular}{C{0.4cm}C{0.5cm}C{0.5cm}|*{2}{C{0.7cm}}||*{6}{C{0.6cm}}||*{6}{C{0.6cm}}}
\hline
\multicolumn{3}{c|}{Methods}  & \multicolumn{2}{c||}{\textbf{STL (Ours)}} & \multicolumn{2}{c|}{\textbf{Linear}}  & \multicolumn{2}{c|}{\textbf{DLinear}} & \multicolumn{2}{c||}{\textbf{ NLinear}} & \multicolumn{2}{c|}{\textbf{Autoformer}}  & \multicolumn{2}{c|}{\textbf{ Informer}} & \multicolumn{2}{c}{\textbf{ FEDformer}} \\
\hline
Data & $T$ &$\tau$  & MSE & MAE & MSE & MAE & MSE & MAE & MSE & MAE & MSE & MAE & MSE  & MAE & MSE & MAE\\
\hline
\multirow{5}{*}{\rot{Electricity}} & \multirow{5}{*}{96} & 24 & \textbf{0.133} & \textbf{0.226} & 0.164 & 0.250 & 0.164 & 0.249 & 0.166 & \underline{0.248} & 0.186 & 0.305 & 0.293 & 0.385 & \underline{0.163} & 0.283 \\
& & 48 & \textbf{0.162} & \textbf{0.253} & 0.195 & 0.275 & 0.195 & 0.274 & 0.199 & \underline{0.273} & 0.187 & 0.303 & 0.307 & 0.396 & \underline{0.175} & 0.292 \\
& & 96 & \textbf{0.171} & \textbf{0.261} & 0.195 & 0.278 & 0.195 & 0.278 & 0.199 & \underline{0.277} & 0.212 & 0.324 & 0.336 & 0.420 & \underline{0.188} & 0.304 \\
& & 192 & \textbf{0.179} & \textbf{0.268} & \underline{0.195} & 0.281 & 0.194 & 0.280 & 0.199 & \underline{0.279} & 0.231 & 0.339 & 0.365 & 0.434 & 0.197 & 0.311 \\
& & 336 & \textbf{0.192} & \textbf{0.284} & \underline{0.208} & 0.297 & 0.207 & 0.297 & 0.213 & \underline{0.295} & 0.250 & 0.356 & 0.389 & 0.455 & 0.212 & 0.327 \\
\hline
\multirow{5}{*}{\rot{Etth1}} & \multirow{5}{*}{96} & 24 & \textbf{0.308} & \textbf{0.351} & 0.320 & 0.362 & 0.312 & \underline{0.354} & 0.317 & 0.358 & 0.403 & 0.437 & 0.546 & 0.521 & \underline{0.314} & 0.380 \\
& & 48 & \textbf{0.342} & \textbf{0.372} & 0.351 & 0.379 & \underline{0.344} & \underline{0.373} & 0.350 & 0.377 & 0.414 & 0.445 & 0.598 & 0.530 & 0.375 & 0.415 \\
& & 96 & \textbf{0.382} & \textbf{0.395} & 0.390 & 0.402 & \underline{0.384} & \underline{0.396} & 0.390 & 0.399 & 0.476 & 0.481 & 0.863 & 0.702 & 0.427 & 0.448 \\
& & 192 & \underline{0.436} & \underline{0.429} & 0.440 & 0.432 & \textbf{0.435} & \textbf{0.427} & 0.441 & 0.428 & 0.540 & 0.536 & 0.983 & 0.764 & 0.427 & 0.448 \\
& & 336 & \underline{0.480} & \underline{0.455} & 0.484 & 0.459 & \textbf{0.478} & \textbf{0.453} & 0.483 & 0.450 & 0.573 & 0.549 & 1.040 & 0.786 & 0.457 & 0.464 \\
\hline
\multirow{5}{*}{\rot{Ettm1}} & \multirow{5}{*}{96} & 24 & \textbf{0.224} & \textbf{0.299} & \underline{0.250} & \underline{0.312} & 0.248 & \underline{0.312} & 0.253 & \underline{0.312} & 0.334 & 0.406 & 0.392 & 0.432 & 0.279 & 0.352 \\
& & 48 & \textbf{0.291} & \textbf{0.341} & \underline{0.315} & \underline{0.352} & 0.320 & 0.357 & 0.326 & 0.357 & 0.389 & 0.432 & 0.518 & 0.503 & 0.344 & 0.394 \\
& & 96 & \textbf{0.328} & \textbf{0.367} & \underline{0.345} & \underline{0.372} & \underline{0.345} & \underline{0.372} & 0.352 & 0.372 & 0.473 & 0.484 & 0.610 & 0.557 & 0.363 & 0.413 \\
& & 192 & \textbf{0.373} & \textbf{0.389} & \underline{0.381} & \underline{0.390} & \underline{0.381} & \underline{0.390} & 0.391 & 0.392 & 0.531 & 0.529 & 0.702 & 0.602 & 0.394 & 0.425 \\
& & 336 & 0.417 & 0.416 & \underline{0.414} & \underline{0.413} & \textbf{0.412} & \textbf{0.412} & 0.424 & \underline{0.413} & 0.521 & 0.511 & 0.912 & 0.698 & 0.455 & 0.460 \\
\hline
\multirow{5}{*}{\rot{Weather}} & \multirow{5}{*}{96} & 24 & \textbf{0.105} & \textbf{0.148} & 0.117 & 0.162 & \underline{0.116} & 0.160 & 0.117 & \underline{0.152} & 0.284 & 0.378 & 0.283 & 0.365 & 0.155 & 0.240 \\
& & 48 & \textbf{0.138} & \textbf{0.192} & 0.159 & 0.217 & \underline{0.152} & 0.202 & 0.157 & \underline{0.198} & 0.417 & 0.452 & 0.382 & 0.454 & 0.192 & 0.279 \\
& & 96 & \textbf{0.171} & \textbf{0.229} & 0.195 & 0.254 & \underline{0.194} & 0.245 & 0.196 & \underline{0.235} & 0.333 & 0.387 & 0.345 & 0.398 & 0.225 & 0.308 \\
& & 192 & \textbf{0.210} & \textbf{0.270} & 0.236 & 0.293 & \underline{0.234} & 0.290 & 0.241 & \underline{0.272} & 0.472 & 0.487 & 0.420 & 0.449 & 0.441 & 0.463 \\
& & 336 & \textbf{0.261} & \underline{0.314} & 0.281 & 0.330 & \underline{0.279} & 0.321 & 0.294 & \textbf{0.308} & 0.460 & 0.452 & 0.579 & 0.546 & 0.510 & 0.502 \\
\hline
\end{tabular}}
\caption{Multivariate time series forecasting errors in MSE and MAE. Observation horizon $T=96$ and prediction horizons $\tau\in\{24,48,96,192,336\}$.The \textbf{top results} are highlighted in \textbf{bold}, and second to best highlighted with \underline{underline}.}
\end{table*}
\begin{table*}[h!]
\centering
{\fontsize{9pt}{10pt}\selectfont
\newcolumntype{C}[1]{>{\centering\arraybackslash}p{#1}}
\begin{tabular}{C{0.4cm}C{0.5cm}C{0.5cm}|*{2}{C{0.7cm}}||*{6}{C{0.6cm}}||*{6}{C{0.6cm}}}
\hline
\multicolumn{3}{c|}{Methods}  & \multicolumn{2}{c||}{\textbf{STL (Ours)}} & \multicolumn{2}{c|}{\textbf{Linear}}  & \multicolumn{2}{c|}{\textbf{DLinear}} & \multicolumn{2}{c||}{\textbf{ NLinear}} & \multicolumn{2}{c|}{\textbf{Autoformer}}  & \multicolumn{2}{c|}{\textbf{ Informer}} & \multicolumn{2}{c}{\textbf{ FEDformer}} \\
\hline
Data & $T$ &$\tau$  & MSE & MAE & MSE & MAE & MSE & MAE & MSE & MAE & MSE & MAE & MSE  & MAE & MSE & MAE\\
\hline
\multirow{5}{*}{\rot{Electricity}} & \multirow{5}{*}{192} & 24 & \textbf{0.100} & \textbf{0.198} & \underline{0.118} & \underline{0.216} & \underline{0.118} & \underline{0.216} & 0.119 & 0.217 & 0.185 & 0.302 & 0.289 & 0.386 & 0.167 & 0.2869 \\
& & 48 & \textbf{0.117} & \textbf{0.214} & \underline{0.135} & \underline{0.232} & \underline{0.135} & \underline{0.232} & 0.136 & 0.233 & 0.186 & 0.303 & 0.318 & 0.410 & 0.177 & 0.2946 \\
& & 96 & \textbf{0.137} & \textbf{0.232} & \underline{0.152} & 0.248 & \underline{0.152} & \underline{0.247} & 0.154 & \underline{0.247} & 0.201 & 0.315 & 0.344 & 0.423 & 0.189 & 0.3047 \\
& & 192 & \textbf{0.154} & \textbf{0.249} & \underline{0.164} & 0.260 & 0.164 & 0.259 & 0.166 & \underline{0.258} & 0.235 & 0.341 & 0.384 & 0.445 & 0.217 & 0.3339 \\
& & 336 &\textbf{0.171} & \textbf{0.268} & \underline{0.180} & 0.277 & \underline{0.180} & 0.277 & 0.183 & \underline{0.275} & 0.258 & 0.360 & 0.382 & 0.447 & 0.227 & 0.3436 \\
\hline
\multirow{5}{*}{\rot{Etth1}} & \multirow{5}{*}{192} & 24 & \textbf{0.315} & \textbf{0.359} & 0.324 & 0.366 & \underline{0.316} & \textbf{0.359} & 0.321 & \underline{0.363} & 0.467 & 0.468 & 0.554 & 0.524 & 0.320 & 0.3886 \\
& & 48 & \textbf{0.343} & \textbf{0.375} & 0.350 & 0.380 & \underline{0.345} & \textbf{0.375} & 0.349 & \underline{0.379} & 0.535 & 0.530 & 0.705 & 0.606 & 0.342 & 0.3981 \\
& & 96 & \underline{0.380} & \underline{0.398} & 0.385 & 0.401 & \textbf{0.379} & \textbf{0.396} & 0.386 & 0.400 & 0.489 & 0.504 & 0.915 & 0.711 & 0.381 & 0.4212 \\
& & 192 & \underline{0.426} & 0.424 & 0.428 & 0.426 & \textbf{0.422} & \textbf{0.421} & 0.428 & \underline{0.423} & 0.527 & 0.516 & 1.019 & 0.772 & 0.432 & 0.4571 \\
& & 336 & 0.465 & 0.452 & \underline{0.460} & 0.447 & \textbf{0.454} & \underline{0.440} & 0.456 & \textbf{0.438} & 0.531 & 0.535 & 1.116 & 0.816 & 0.443 & 0.4622 \\
\hline
\multirow{5}{*}{\rot{Ettm1}} & \multirow{5}{*}{192} & 24 & \textbf{0.199} & \textbf{0.282} & \underline{0.218} & 0.290 & \underline{0.218} & \underline{0.288} & 0.220 & 0.289 & 0.477 & 0.492 & 0.349 & 0.404 & 0.299 & 0.373 \\
& & 48 & \textbf{0.262} & \textbf{0.324} & 0.278 & 0.328 & \underline{0.277} & \underline{0.327} & 0.281 & 0.329 & 0.422 & 0.454 & 0.479 & 0.489 & 0.359 & 0.4099 \\
& & 96 & \textbf{0.296} & 0.348 & \underline{0.306} & \textbf{0.345} & \underline{0.306} & \textbf{0.345} & 0.309 & \underline{0.346} & 0.456 & 0.482 & 0.549 & 0.530 & 0.350 & 0.4069 \\
& & 192 & \textbf{0.336} & \textbf{0.367} & 0.342 & \textbf{0.367} & \underline{0.341} & \textbf{0.367} & 0.345 & \textbf{0.367} & 0.427 & 0.461 & 0.763 & 0.642 & 0.382 & \underline{0.423} \\
& & 336 & \textbf{0.376} & \textbf{0.407} & \underline{0.377} & \underline{0.387} & \textbf{0.376} & 0.388 & 0.382 & 0.388 & 0.509 & 0.498 & 0.982 & 0.736 & 0.415 & 0.4433 \\
\hline
\multirow{5}{*}{\rot{Weather}} & \multirow{5}{*}{192} & 24 & \textbf{0.094} & \textbf{0.134} & 0.109 & 0.157 & \underline{0.107} & 0.156 & 0.109 & \underline{0.150} & 0.253 & 0.341 & 0.228 & 0.317 & 0.189 & 0.276 \\
& & 48 & \textbf{0.124} &\textbf{0.177} & 0.150 & 0.215 & 0.146 & 0.199 & \underline{0.145} & \underline{0.189} & 0.309 & 0.390 & 0.335 & 0.409 & 0.227 & 0.3171 \\
& & 96 & \textbf{0.157} & \textbf{0.217} & 0.185 & 0.244 & \underline{0.184} & 0.243 & 0.186 & \underline{0.228} & 0.340 & 0.427 & 0.395 & 0.431 & 0.252 & 0.33 \\
& & 192 & \textbf{0.200} & \textbf{0.259} & \underline{0.224} & 0.281 & 0.226 & 0.280 & 0.230 & \underline{0.265} & 0.423 & 0.458 & 0.487 & 0.495 & 0.287 & 0.3546 \\
& & 336 & \textbf{0.254} & \underline{0.306} & \underline{0.269} & 0.316 & \underline{0.269} & 0.309 & 0.279 & \textbf{0.302} & 0.475 & 0.488 & 0.603 & 0.572 & 0.333 & 0.3788 \\
\hline
\end{tabular}}
\caption{Multivariate time series forecasting errors in MSE and MAE. Observation horizon $T=192$ and prediction horizons $\tau\in\{24,48,96,192,336\}$.The \textbf{top results} are highlighted in \textbf{bold}, and second to best highlighted with \underline{underline}.}
\end{table*}
\begin{table*}[h!]
\centering
{\fontsize{9pt}{10pt}\selectfont
\newcolumntype{C}[1]{>{\centering\arraybackslash}p{#1}}
\begin{tabular}{C{0.4cm}C{0.5cm}C{0.5cm}|*{2}{C{0.7cm}}||*{6}{C{0.6cm}}||*{6}{C{0.6cm}}}
\hline
\multicolumn{3}{c|}{Methods}  & \multicolumn{2}{c||}{\textbf{STL (Ours)}} & \multicolumn{2}{c|}{\textbf{Linear}}  & \multicolumn{2}{c|}{\textbf{DLinear}} & \multicolumn{2}{c||}{\textbf{ NLinear}} & \multicolumn{2}{c|}{\textbf{Autoformer}}  & \multicolumn{2}{c|}{\textbf{ Informer}} & \multicolumn{2}{c}{\textbf{ FEDformer}} \\
\hline
Data & $T$ &$\tau$  & MSE & MAE & MSE & MAE & MSE & MAE & MSE & MAE & MSE & MAE & MSE  & MAE & MSE & MAE\\
\hline
\multirow{5}{*}{\rot{Electricity}} & \multirow{5}{*}{336} & 24 & \textbf{0.097} & \textbf{0.195} & \underline{0.110} & 0.210 & \underline{0.110} & \underline{0.209} & 0.111 & 0.210 & 0.196 & 0.317 & 0.312 & 0.403 & 0.176 & 0.296 \\
& & 48 & \textbf{0.112} & \textbf{0.209} & \underline{0.125} & \underline{0.224} & \underline{0.125} & \underline{0.224} & 0.126 & \underline{0.224} & 0.205 & 0.323 & 0.314 & 0.404 & 0.185 & 0.302 \\
& & 96 & \textbf{0.131} & \textbf{0.227} & \underline{0.140} & \underline{0.238} & \underline{0.140} & \underline{0.238} & 0.142 & \underline{0.238} & 0.210 & 0.325 & 0.337 & 0.419 & 0.205 & 0.322 \\
& & 192 & \textbf{0.150} & \textbf{0.246} & \underline{0.154} & 0.251 & \underline{0.154} & 0.251 & 0.156 & \underline{0.250} & 0.225 & 0.341 & 0.377 & 0.443 & 0.212 & 0.330 \\
& & 336 & \textbf{0.167} & \textbf{0.262} & \underline{0.170} & 0.269 & \underline{0.170} & 0.268 & 0.173 & \underline{0.267} & 0.233 & 0.344 & 0.383 & 0.448 & 0.243 & 0.357 \\
\hline
\multirow{5}{*}{\rot{Etth1}} & \multirow{5}{*}{336} & 24 & \textbf{0.319} & \textbf{0.363} & 0.327 & 0.369 & \underline{0.320} & \textbf{0.363} & 0.325 & \underline{0.368} & 0.528 & 0.527 & 0.691 & 0.596 & 0.334 & 0.404 \\
& & 48 & \underline{0.345} & \underline{0.378} & 0.350 & 0.382 & \textbf{0.344} & \textbf{0.377} & 0.350 & 0.382 & 0.523 & 0.508 & 0.819 & 0.664 & 0.353 & 0.410 \\
& & 96 & \textbf{0.372} & \underline{0.394} & 0.377 & 0.398 & \textbf{0.372} & \textbf{0.393} & \underline{0.376} & 0.398 & 0.506 & 0.491 & 0.977 & 0.735 & 0.384 & 0.427 \\
& & 192 & \textbf{0.404} & \textbf{0.413} & 0.410 & \underline{0.418} & \underline{0.405} & \textbf{0.413} & 0.411 & \underline{0.418} & 0.509 & 0.507 & 1.053 & 0.762 & 0.429 & 0.455 \\
& & 336 & \textbf{0.433} & 0.438 & 0.439 & 0.440 & \underline{0.434} & \underline{0.434} & 0.435 & \textbf{0.433} & 0.526 & 0.529 & 1.135 & 0.797 & 0.456 & 0.478 \\
\hline
\multirow{5}{*}{\rot{Ettm1}} & \multirow{5}{*}{336} & 24 & \textbf{0.194} & \textbf{0.281} & \underline{0.211} & \underline{0.284} & \underline{0.211} & \underline{0.284} & 0.212 & \underline{0.284} & 0.331 & 0.401 & 0.401 & 0.441 & 0.319 & 0.390 \\
& & 48 & \textbf{0.264} & 0.328 & \underline{0.271} & \underline{0.326} & 0.273 & 0.328 & 0.272 & \textbf{0.325} & 0.488 & 0.489 & 0.509 & 0.490 & 0.380 & 0.429 \\
& & 96 & \textbf{0.291} & \textbf{0.342} & \underline{0.300} & \underline{0.343} & 0.299 & \underline{0.343} & 0.303 & 0.345 & 0.546 & 0.502 & 0.639 & 0.586 & 0.350 & 0.410 \\
& & 192 & \textbf{0.330} & 0.366 & 0.335 & \textbf{0.364} & \underline{0.334} & \textbf{0.364} & 0.338 & \underline{0.365} & 0.500 & 0.502 & 0.805 & 0.660 & 0.379 & 0.425 \\
& & 336 & \underline{0.371} & 0.395 & \textbf{0.369} & \underline{0.385} & \textbf{0.369} & \underline{0.385} & \underline{0.371} & \textbf{0.384} & 0.511 & 0.496 & 1.100 & 0.793 & 0.432 & 0.451 \\
\hline
\multirow{5}{*}{\rot{Weather}} & \multirow{5}{*}{336} & 24 & \textbf{0.092} & \textbf{0.130} & \underline{0.103} & 0.148 & \underline{0.103} & 0.149 & \underline{0.103} & \textbf{0.141} & 0.362 & 0.426 & 0.237 & 0.325 & 0.217 & 0.302 \\
& & 48 & \textbf{0.119} & \textbf{0.167} & 0.138 & 0.197 & 0.137 & 0.194 & \underline{0.136} & \underline{0.183} & 0.362 & 0.426 & 0.313 & 0.383 & 0.274 & 0.359 \\
& & 96 & \textbf{0.153} & \textbf{0.211} & 0.175 & 0.235 & \underline{0.173} & 0.231 & 0.176 & \underline{0.224} & 0.370 & 0.428 & 0.452 & 0.478 & 0.277 & 0.354 \\
& & 192 & \textbf{0.198} & \textbf{0.255} & \underline{0.215} & 0.272 & 0.216 & 0.266 & 0.217 & \underline{0.260} & 0.452 & 0.470 & 0.630 & 0.570 & 0.280 & 0.350 \\
& & 336 & \textbf{0.253} & \underline{0.304} & \underline{0.260} & 0.309 & 0.262 & \underline{0.304} & 0.264 & \textbf{0.294} & 0.475 & 0.481 & 0.642 & 0.568 & 0.344 & 0.388 \\
\hline
\end{tabular}}
\caption{Multivariate time series forecasting errors in MSE and MAE. Observation horizon $T=336$ and prediction horizons $\tau\in\{24,48,96,192,336\}$.The \textbf{top results} are highlighted in \textbf{bold}, and second to best highlighted with \underline{underline}.}
\end{table*}
\begin{table*}[h!]
\centering
{\fontsize{9pt}{10pt}\selectfont
\newcolumntype{C}[1]{>{\centering\arraybackslash}p{#1}}
\begin{tabular}{C{0.4cm}C{0.5cm}C{0.5cm}|*{2}{C{0.7cm}}||*{6}{C{0.6cm}}||*{6}{C{0.6cm}}}
\hline
\multicolumn{3}{c|}{Methods}  & \multicolumn{2}{c||}{\textbf{STL (Ours)}} & \multicolumn{2}{c|}{\textbf{Linear}}  & \multicolumn{2}{c|}{\textbf{DLinear}} & \multicolumn{2}{c||}{\textbf{ NLinear}} & \multicolumn{2}{c|}{\textbf{Autoformer}}  & \multicolumn{2}{c|}{\textbf{ Informer}} & \multicolumn{2}{c}{\textbf{ FEDformer}} \\
\hline
Data & $T$ &$\tau$  & MSE & MAE & MSE & MAE & MSE & MAE & MSE & MAE & MSE & MAE & MSE  & MAE & MSE & MAE\\
\hline
\multirow{5}{*}{\rot{Electricity}} & \multirow{5}{*}{504}  & 24 & \textbf{0.094} & \textbf{0.192} & \underline{0.106} & \underline{0.205} & \underline{0.106} & \underline{0.205} & 0.107 & 0.206 & 0.207 & 0.327 & 0.311 & 0.402 & 0.182 & 0.301 \\
& & 48 & \textbf{0.110} & \textbf{0.208} & \underline{0.121} & 0.220 & \underline{0.121} & \underline{0.219} & 0.122 & 0.220 & 0.212 & 0.330 & 0.333 & 0.420 & 0.206 & 0.320 \\
& & 96 & \textbf{0.129} & \textbf{0.225} & \underline{0.136} & 0.234 & \underline{0.136} & \underline{0.233} & 0.138 & 0.234 & 0.223 & 0.337 & 0.338 & 0.418 & 0.212 & 0.327\\
& & 192 & \textbf{0.150} & \underline{0.247} & \textbf{0.150} & \underline{0.247} & \textbf{0.150} & \underline{0.247} & \underline{0.152} & \textbf{0.246} & 0.227 & 0.340 & 0.367 & 0.436 & 0.216 & 0.330\\
& & 336 & \textbf{0.163} & \textbf{0.260} & \underline{0.165} & 0.264 & \underline{0.165} & 0.265 & 0.168 & \underline{0.263} & 0.260 & 0.369 & 0.386 & 0.450 & 0.231 & 0.345\\
\hline
\multirow{5}{*}{\rot{Etth1}} & \multirow{5}{*}{504}  & 24 & \underline{0.314} & \textbf{0.362} & 0.320 & 0.368 & \textbf{0.313} & \textbf{0.362} & 0.317 & \underline{0.366} & 0.529 & 0.545 & 0.794 & 0.626 & 0.345 & 0.416 \\
& & 48 & \textbf{0.340} & \textbf{0.376} & 0.344 & 0.381 & \textbf{0.340} & \textbf{0.376} & \underline{0.342} & \underline{0.380} & 0.538 & 0.542 & 0.875 & 0.667 & 0.371 & 0.429 \\
& & 96 & \textbf{0.370} & \textbf{0.394} & 0.373 & 0.398 & \textbf{0.370} & \textbf{0.394} & \underline{0.371} & \underline{0.397} & 0.500 & 0.520 & 1.029 & 0.746 & 0.405 & 0.449\\
& & 192 & 0.410 & 0.422 & 0.409 & 0.422 & \textbf{0.404} & \textbf{0.416} & \underline{0.406} & \underline{0.417} & 0.541 & 0.530 & 1.195 & 0.812 & 0.435 & 0.465 \\
& & 336 & 0.446 & 0.451 & \underline{0.439} & 0.442 & \underline{0.440} & 0.443 & \textbf{0.435} & \textbf{0.436} & 0.599 & 0.572 & 1.239 & 0.853 & 0.451 & 0.475 \\
\hline
\multirow{ 5}{*}{\rot{Ettm1}} & \multirow{5}{*}{504}  & 24 & \textbf{0.197} & \textbf{0.280} & \underline{0.215} & \underline{0.286} & \underline{0.215} & \underline{0.286} & 0.216 & 0.287 & 0.360 & 0.418 & 0.408 & 0.443 & 0.348 & 0.407 \\
& & 48 & \textbf{0.270} & 0.331 & \underline{0.276} & \underline{0.330} & \underline{0.276} & \textbf{0.328} & 0.279 & 0.331 & 0.505 & 0.500 & 0.515 & 0.516 & 0.391 & 0.439 \\
& & 96 & \textbf{0.294} & \textbf{0.344} & \underline{0.303} & \underline{0.346} & 0.309 & 0.355 & 0.305 & 0.347 & 0.543 & 0.515 & 0.642 & 0.579 & 0.384  & 0.432\\
& & 192 & \textbf{0.331} & \textbf{0.366} & 0.337 & \underline{0.368} & \underline{0.335} & \textbf{0.366} & 0.341 & 0.369 & 0.506 & 0.511 & 0.915 & 0.729 & 0.390 & 0.435\\
& & 336 & 0.374 & 0.398 & 0.368 & \underline{0.386} & \textbf{0.366} & \textbf{0.384} & \underline{0.367} & \textbf{0.384} & 0.493 & 0.507 & 1.047 & 0.763 & 0.419 & 0.450 \\
\hline
\multirow{5}{*}{\rot{Weather}} & \multirow{5}{*}{504} & 24 & \textbf{0.092} & \textbf{0.129} & 0.102 & 0.146 & \underline{0.101} & 0.147 & 0.102 & \underline{0.145} & 0.367 & 0.435 & 0.208 & 0.296 & 0.262 & 0.344\\
& & 48 & \textbf{0.117} & \textbf{0.165} & 0.134 & 0.189 & \underline{0.133} & 0.188 & 0.134 & \underline{0.187} & 0.367 & 0.426 & 0.389 & 0.441 & 0.296 & 0.368\\
& & 96 & \textbf{0.149} & \textbf{0.206} & \underline{0.169} & 0.226 & 0.170 & 0.227 & 0.170 & \underline{0.222} & 0.437 & 0.466 & 0.459 & 0.475 & 0.311 & 0.371\\
& & 192 & \textbf{0.195} & \textbf{0.253} & 0.214 & 0.270 & \underline{0.211} & 0.264 & 0.216 & \underline{0.263} & 0.473 & 0.487 & 0.567 & 0.552 & 0.332 & 0.390\\
& & 336 & \textbf{0.250} & \underline{0.300} & \underline{0.257} & 0.304 & 0.258 & 0.306 & 0.262 & \textbf{0.296} & 0.478 & 0.498 & 0.765 & 0.646 & 0.571 & 0.545\\
\hline
\end{tabular}}
\caption{Multivariate time series forecasting errors in MSE and MAE. Observation horizon $T=504$ and prediction horizons $\tau\in\{24,48,96,192,336\}$.The \textbf{top results} are highlighted in \textbf{bold}, and second to best highlighted with \underline{underline}.}
\end{table*}

\end{document}